\documentclass[a4paper,fleqn]{cas-dc}
\usepackage{amsmath,amsfonts}
\usepackage{algorithmic}
\usepackage{algorithm}
\usepackage{array}
\usepackage{graphicx}
\usepackage{cite}
\usepackage{braket}
\usepackage{hyperref} 
\usepackage{amsthm}
\usepackage{textcomp}
\usepackage{stfloats}
\usepackage{url}
\usepackage{verbatim}
\usepackage[caption=false,font=normalsize,labelfont=sf,textfont=sf]{subfig}

\usepackage{natbib}

\def\tsc#1{\csdef{#1}{\textsc{\lowercase{#1}}\xspace}}
\tsc{WGM}
\tsc{QE}
\tsc{EP}
\tsc{PMS}
\tsc{BEC}
\tsc{DE}

\newtheorem{theorem}{Theorem}
\newtheorem{lemma}{Lemma}

\newtheorem{assumption}{Assumption}

\newcommand{\textremovedforicml}[1]{}

\newcommand{\E}{\mathbb{E}}

\newcommand{\Eb}[2]{\E_{#1}\left[#2\right]}

\newcommand{\cS}{\mathcal{S}}
\newcommand{\cA}{\mathcal{A}}
\newcommand{\cP}{P}
\newcommand{\Real}{\mathbb{R}}

\newcommand{\rhopi}{\rho_{\pi}}

\newcommand{\product}[1]{\langle #1\rangle_{\mathcal{F}}}
\newcommand{\kl}[2]{D_{\rm KL}(#1 \ \| \ #2)}

\newcommand{\half}{\frac{1}{2}}

\newcommand{\Api}{A^{\boldsymbol{\pi}}}

\newcommand{\orcid}[1]{\href{https://orcid.org/#1}{\includegraphics[width=10pt]{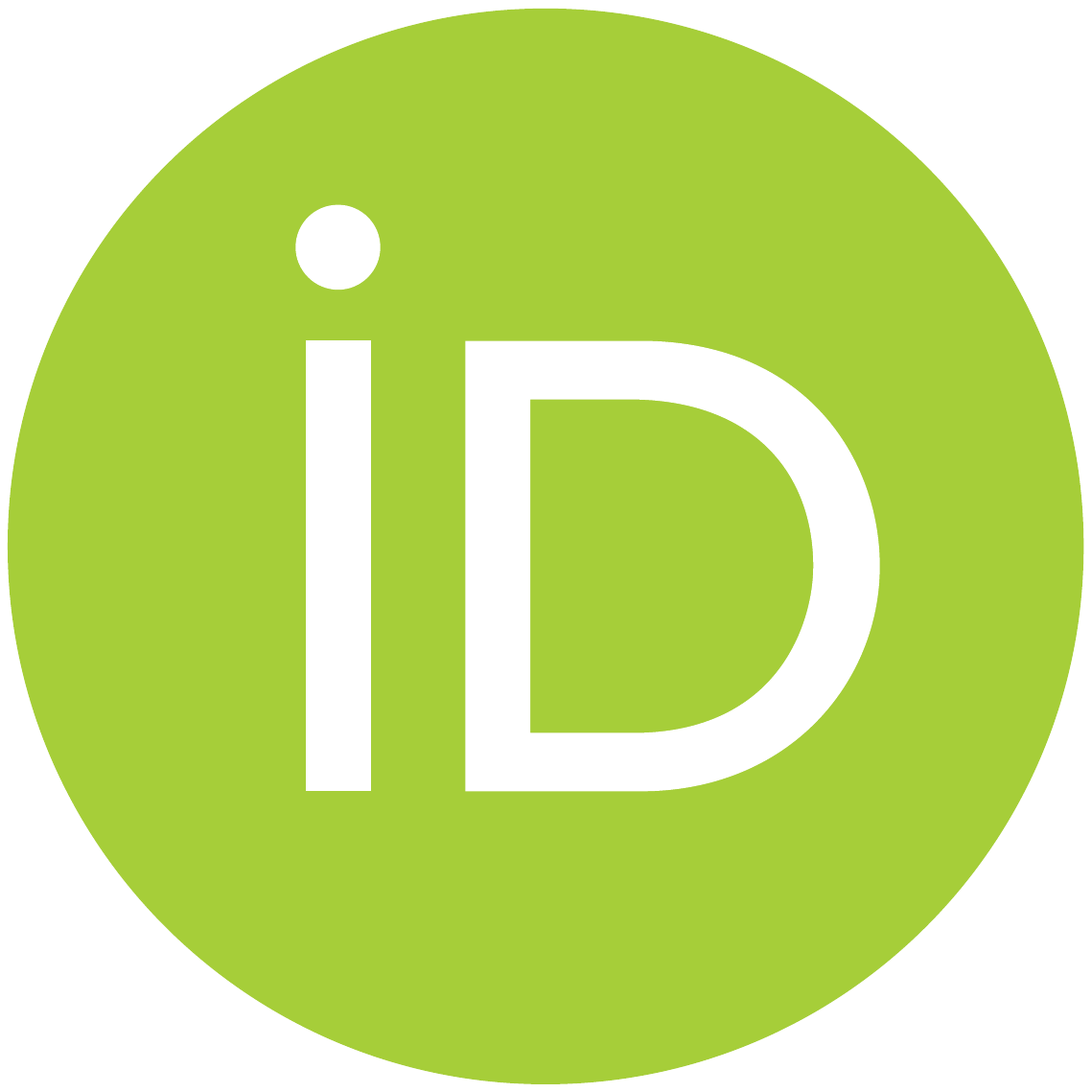}}}

\begin{document}
\let\WriteBookmarks\relax
\def\floatpagepagefraction{1}
\def\textpagefraction{.001}
\shorttitle{Proximal Policy Optimization with Correntropy Induced Metric}
\shortauthors{Y. Guo et~al.}
\title[mode = title]{PPO-CIM: Proximal Policy Optimization with Correntropy Induced Metric}\tnotemark[1]
\tnotetext[1]{This paper is the result of the research project funded by the National Science Foundation.}
\author[a]{Yunxiao Guo}[orcid=0000-0003-2078-0393]
\ead{guoyunxiao.nudt@hotmail.com}
\author[b,c]{Dan Xu}
\ead{xudan@nudt.edu.cn}
\author[a]{Xiaojun Duan}
\ead{xjduan@nudt.edu.cn}
\author[a]{Kaiyuan Feng}
\ead{fengkaiyuan@nudt.edu.cn}
\author[a]{Maochu Li}
\ead{limaochu@alumni.nudt.edu.cn}
\author[a]{Xiaying Ma}
\ead{maxiaying@nudt.edu.cn}
\author[a]{Han Long}[orcid=0000-0003-2947-5758]\ead{longhan@nudt.edu.cn}\cormark[1]
\cortext[1]{Corresponding author: Han Long}
\affiliation[a]{organization={College of Science, National University of Defense Technology},city={Changsha},
citysep={}, postcode={410073}, state={Hunan},country={China}}
\affiliation[b]{organization={School of System Science and Engineering, Sun-Yat-Sen University},city={Guangzhou},
citysep={}, postcode={510006}, state={Guangdong},country={China}}
\affiliation[c]{organization={College of System Engineering, National University of Defense Technology},city={Changsha},
citysep={}, postcode={410073}, state={Hunan},country={China}}


\maketitle
\begin{abstract}
As a popular Deep Reinforcement Learning (DRL) algorithm, Proximal Policy Optimization (PPO) has demonstrated remarkable efficacy in numerous complex tasks. According to the penalty mechanism in a surrogate, PPO can be classified into PPO with KL divergence (PPO-KL) and PPO with Clip (PPO-Clip). In this paper, we analyze the impact of asymmetry in KL divergence on PPO-KL and highlight that when this asymmetry is pronounced, it will misguide the improvement of the surrogate. To address this issue, we represent the PPO-KL in inner product form and demonstrate that the KL divergence is a Correntropy Induced Metric (CIM) in Euclidean space. Subsequently, we extend the PPO-KL to the Reproducing Kernel Hilbert Space (RKHS), redefine the inner products with RKHS, and propose the PPO-CIM algorithm. Moreover, this paper states that the PPO-CIM algorithm has a lower computation cost in policy gradient and proves that PPO-CIM can guarantee the new policy is within the trust region while the kernel satisfies some conditions. Finally, we design experiments based on six Mujoco continuous-action tasks to validate the proposed algorithm. The experimental results validate that the asymmetry of KL divergence can affect the policy improvement of PPO-KL and show that the PPO-CIM can perform better than both PPO-KL and PPO-Clip in most tasks.

\end{abstract}

\section{Introduction}
Deep reinforcement learning (DRL) aims to obtain the optimal policy for agents interacting with the environment by using neural networks~\citep{DRL}. The Proximal Policy Optimization (PPO) algorithm is a branch of the policy gradient-based DRL methods~\citep{ppo}. As one of the most popular algorithms, PPO was applied in various areas successfully, such as Large Language Model (LLM) optimization~\citep{HFRL, HFRL3, HFRL2}, task allocation~\citep{PPOKBS, PPO-TA}, Unmanned Aerial Vehicle (UAV) control~\citep{PPO-Exp, UAV1, UAV2} and autonomous driving ~\citep{drive, drive3, drive2}.

Policy gradient-based reinforcement learning methods aim to adjust the policy parameters through the gradient information to optimize the agent behavior~\citep{PG}. It also faces some problems: On the one hand, it is hard to converge on locally optimal policy while running into a plateau area~\citep{npg}. On the other hand, it can not guarantee the future reward is monotone increasing after a update~\citep{cpi}. To ensure policy optimality, \cite{npg} proposed the Natural Policy Gradient (NPG) method and proved the NPG can provide the policy update with a greedy optimal action rather than a better action. To guarantee the policy improvement, \citep{cpi} defined the updated policy's performance as the old policy adds a non-negative increase, then performed a theoretical analysis to guarantee the policy improvement. To update and evaluate the policy simultaneously, \cite{trpo} introduced the Actor-Critic framework into the NPG, which utilizes an actor to update the policy through policy gradient and a critic to evaluate the policy. Furthermore, to ensure policy improvement, TRPO considers the Kullback–Leibler divergence (KL-divergence) between new and old policies as the trust region constraint to restrict the updating range.

However, as an optimization method with trust region constraints, TRPO also struggles with learning efficiency. Although sampling and approximate techniques were tokens, the TRPO still performs pool on the large-scale learning tasks~\citep{ppo}. To accelerate the learning process, \cite{ppo} further proposed the PPO with KL divergence (PPO-KL) algorithm, which transfers the trust region optimization problem to a non-constrain optimization problem by adding the trust region constraint as a penalty on the surrogate. Nevertheless, the PPO-KL still needs to compute a large amount of Fisher information matrix to approximate the KL divergence, which also has a high computational cost. Therefore, by replacing the KL divergence with a Clip function, the PPO with Clip (PPO-Clip) algorithm was proposed~\citep{ppo}. With a computable clip function, PPO-Clip can handle large-scale tasks with fewer calculations and perform surprisingly in most DRL tasks~\citep{MAPPO}. 

Because the PPO-Clip lacks theoretical support~\citep{NPPOClip}, most of the current improved methods based on PPO-Clip are focused on improving their learning framework~\citep{Sys6, PPOARC} or balancing their exploration and exploitation~\citep{PPO-Exp, ACPPO}. Specifically, from the perspective of the learning framework, \cite{Sys6} designed a novel Actor-Critic framework by introducing the policy information into the process of the value function update and proposed the PPO with Policy Feedback (PPO-PF) algorithm. \cite{AMPPO} designed an anti-martingale reinforcement learning framework to select the samples efficiently and proposed the Anti-Martingale PPO (AMPPO) algorithm that combined the anti-martingale RL framework with PPO. \cite{PPOARC} introduced the advantage reuse technique and parallel competitive optimization into PPO and proposed the PPO With Advantage Reuse Competition (PPO-ARC) algorithm. From the perspective of balancing the exploration and exploitation, \cite{ACPPO} and \cite{PPO-Exp} pointed out that PPO-Clip's fixed setting of clipping bound restricts their exploration ability. To overcome this weakness, \cite{ACPPO} introduced the adaptive clipping mechanism to adjust the clipping bound and proposed the PPO-$\lambda$ algorithm. \cite{PPOCMA} dynamically expanded and contracted the exploration variance of PPO and proposed PPO with the Covariance Matrix Adaptation (PPO-CMA) algorithm. \cite{PPO-RPD} pointed out the unbalanced regularization problem in PPO-Clip and proposed the PPO with Relative Pearson Divergence (PPO-RPE) algorithm that used the relative Pearson divergence as the constraint to alleviate it. To improve the exploration efficiency, \cite{PPO-Exp} proposed PPO with $\epsilon$ Exploration (PPO-Exp) algorithm, which decomposes the advantage function as the regular and exploration parts, then utilized exploration part to adjust the policy exploration. \cite{PBPPO} used a multi-armed bandit algorithm to indicate the adjustment of the clipping bound in each epoch and proposed the Preference-based PPO (Pb-PPO) algorithms.

DRL agents in specific tasks require high reliability~\citep{HLong2}, so the weakness in theoretical substantiation and interpretability will restrict the improvement of the PPO-Clip algorithm. To compensate for the lack of theoretical substantiation of PPO-Clip, some studies have tried to establish the connection between PPO-Clip and previous algorithms in recent years~\citep{APPO, NPPOClip}. \cite{APPO} studied the relationship between TRPO and PPO-Clip, analyzed the effect of clip operation on the PPO-Clip algorithm, and proposed a novel first-order policy gradient algorithm called Authentic Boundary PPO (ABPPO) algorithm based on the authentic boundary setting rule. \cite{NPPOClip} generalized the PPO-Clip to encompass a broader range of variants by connecting PPO-Clip and Hinge loss, then gave the first global convergence results. From another point of view, as the algorithm that directly evolved by TRPO, PPO-KL is a more theoretical and interpretable algorithm than PPO-Clip. Compared with PPO-Clip, the PPO-KL is different in that it uses the KL divergence to restrict policy updates. KL-divergence is an asymmetric and unbounded function to measure the difference between two distributions, a special case of f-divergence~\citep{fDiv}. \cite{TRPO2} pointed out that KL divergence is unsuitable for TRPO. \cite{KL2} demonstrate that KL divergence is not a reliable constraint to control the behavior of an advanced RL agent. A similar problem of KL divergence also appears in other fields of machine learning. In WGAN, the KL divergence was replaced by the Wasserstein distance, which significantly improved the generation's performance~\citep{WGAN}.

In this paper, we analyze the effect of the asymmetry of KL divergence, represent the PPO-KL algorithm as inner product form, and propose a new algorithm named PPO with Corresponding Induced Metric (PPO-CIM). The most related to our work is AlphaPPO. \cite{PPOAD} reformulated the PPO as a linearly combined form and proposed a novel algorithm named AhphaPPO, which utilizes the alpha divergence to measure the policy difference in the training stage.

The main contributions of this paper are shown as follows:

\begin{enumerate}
\item We analyze the asymmetry of KL divergence's influence on PPO-KL and found that policy improvement can not be guaranteed when the asymmetry is significant enough. Specifically, an inequality is given through theoretical derivation. When it holds, the asymmetry of KL divergence will misguide the surrogate improvement.
\item We expand the PPO-KL as inner product form and discover the KL penalty is a Correntropy Induced Metric in Euclidean Space. Therefore, we extend PPO-KL to PPO-CIM by introducing a Reproducing Kernel Hilbert Space (RKHS) to redefine the inner products in PPO-KL. 

\item We give two conditions to associate the selection of RKHS: the first indicates whether CIM will reduce the computational complexity; the second indicates whether the PPO-CIM can ensure the policy update within the trust region. Furthermore, we give two approximations, PPO-CIM-1 and PPO-CIM-2, to reduce computation and accelerate the learning process.

\end{enumerate}

This paper is divided into five sections. The preliminaries are followed by the second section, which illustrates the foundation of reinforcement learning and briefly introduces PPO and CIM. The third section analyzes the weakness of the PPO-KL algorithm and proposes PPO-CIM. The fourth section is the experiment section, and the fifth section concludes.

\section{Preliminaries}
\subsection{Markov Decision Process}
A reinforcement learning task that satisfies the Markov property is called Markov Decision Process (MDP)~\citep{kkd, 1}. A MDP is defined by the tuple $(\cS, \cA, \cP, r, d_0, \gamma)$ where $s\in\cS$ is the state, $a\in\cA$ is the action, $\cP: \cS \times \cA \times \cS \rightarrow \Real$ is the transition probability distribution, $r: \cS \times \cA \rightarrow \Real$ is the reward function, $d_0: \cS \rightarrow \Real$ is the initial state distribution, and $\gamma \in (0,1)$ is the discount factor.

In some tasks, the policy can be modeled as a distribution $\pi(a_t|s_t)$, representing the action $a_t$'s selected probability under the state $s_t$. Denote the state value function $V_\pi$ and the state-action value function $Q_\pi$ as follows, respectively:
\begin{align}
Q_{\pi}(s_t,a_t) =& \Eb{\pi}{\sum_{k=0}^\infty \gamma^k r_{t+k}\vert(s=s_t,a=a_t)},\label{return}\\
V_{\pi}(s_t) =& \Eb{\pi}{\sum_{k=0}^\infty \gamma^k r_{t+k}\vert(s=s_t)}.
\end{align}
Direct maximizes the state-action value function will introduce high variance~\citep{PG}. To alleviate it, the state value function is introduced as the baseline:
\begin{align}
\Api(s_t,a_t) = Q_{\pi}(s_t,a_t) - V_\pi(s_t).\label{return}
\end{align}
Therefore, the target of the policy based reinforcement learning is to adjust the policy to maximize the advantage function.

\subsection{Proximal Policy Optimization with KL divergence}
Proximal Policy Optimization (PPO) is a model-free DRL algorithm that performs well in discrete and continuous tasks~\citep{ppo}. Consider a continuous action space task with a normal distributional policy parameterized by the actor network parameters $\boldsymbol{\theta}$~\citep{TPPO, MulN, mine}:
\begin{align}
\boldsymbol{\pi}_{\boldsymbol{\theta}}(\boldsymbol{a}|\boldsymbol{s}_t)\sim \mathcal{N}(\boldsymbol{a}|f_{\boldsymbol{\theta}}^\mu(\boldsymbol{s}_t),f_{\boldsymbol{\theta}}^\Sigma(\boldsymbol{s}_t))
\end{align}
It optimizes the non-constraint surrogate objective function to obtain the optimal control policy of tasks. The PPO algorithms can be divided into two categories: PPO with KL-divergence (PPO-KL) and PPO with Clip function (PPO-Clip). In this paper, we mainly discuss the PPO-KL algorithm. The surrogate objective of PPO-KL can be represented as follows:

\begin{align}
&\mathcal{L}^{\text{KL}}(\boldsymbol{\pi}_{\boldsymbol{\theta}},\boldsymbol{\pi}_{\boldsymbol{\theta}_{\text{old}}})\nonumber \\
=&\Eb{\boldsymbol{\pi}_{\boldsymbol{\theta}_{\text{old}}}}{\frac{\boldsymbol{\pi}_{\boldsymbol{\theta}}(\boldsymbol{a}|\boldsymbol{s})}{\boldsymbol{\pi}_{\boldsymbol{\theta}_{\text{old}}}(\boldsymbol{a}|\boldsymbol{s})} \Api(\boldsymbol{a},\boldsymbol{s})-\beta \cdot \kl{\boldsymbol{\pi}_{\boldsymbol{\theta}_{\text{old}}}}{\boldsymbol{\pi}_{\boldsymbol{\theta}}} }.\label{eq:KLso}
\end{align}

where the KL-divergence $D_{\text{KL}}$ is introduced to measure the distance between old and new policies. It can be represented as follows:
\begin{align}
&D_{\text{KL}}(\boldsymbol{\pi}_{\boldsymbol{\theta}}\|\boldsymbol{\pi}_{\boldsymbol{\theta}_{\text{old}}})=\int_{\boldsymbol{a}} \boldsymbol{\pi}_{\boldsymbol{\theta}}(\boldsymbol{a}|\boldsymbol{s})\ln\frac{\boldsymbol{\pi}_{\boldsymbol{\theta}}(\boldsymbol{a}|\boldsymbol{s})}{\boldsymbol{\pi}_{\boldsymbol{\theta}_{\text{old}}}(\boldsymbol{a}|\boldsymbol{s})}d\boldsymbol{a}.
\end{align}
the $\beta$ in Eq.(\ref{eq:KLso}) is an adaptive coefficient, it would adjust the weight of KL divergence according to the following rule:
\begin{align}
\beta =\left\{\begin{matrix}
\beta/2, &\text{if } \mathbb{E}[D_{\rm KL}(\boldsymbol{\pi}_{\boldsymbol{\theta}_{\rm old}},\boldsymbol{\pi}_{\boldsymbol{\theta}})]<D_{\rm KL}^{tar}/1.5\\
1.5\beta, &\text{if } \mathbb{E}[D_{\rm KL}(\boldsymbol{\pi}_{\boldsymbol{\theta}_{\rm old}},\boldsymbol{\pi}_{\boldsymbol{\theta}})]>1.5D_{\rm KL}^{tar}\\
\beta, &\text{otherwise}
\end{matrix}\right.
\end{align} 
where the $D_{\rm KL}^{tar}$ is the target KL-divergence, which indicate whether the KL divergence is too large or small.

\subsection{Correntropy and It's Induced Metric}
\cite{corr} proposed a similarity measure named Correntropy. According to Renyi’s quadratic entropy, correntropy is a generalized similarity measure between two arbitrary scalar random variables $p$ and $q$ defined as:
\begin{align}
&V_\sigma(p,q)=\Eb{}{\kappa_\sigma(p-q)}, \label{eq:cor}
\end{align}
where $\kappa_\sigma(\cdot)$ is a kernel function, $\sigma$ is the kernel size. The Gaussian kernel is one of the most popular kernels that can be represented as follows:
\begin{align}
    &\kappa_\sigma(p_i-q_i)=\frac{1}{\sqrt{2\pi\sigma}}\exp\bigg(\frac{-\|p_i-q_i\|^2}{2\sigma^2}\bigg)\label{gkernel}
\end{align}
From the perspective of Reproducing Kernel Hilbert Space (RKHS), the kernel function can be represented as the inner product format with a nonlinear mapping  $\Phi$: 
\begin{align}
&\kappa_\sigma(p-q)=\product{\Phi(p),\Phi(q)}\label{kappa}
\end{align}
where $\mathcal{F}$ represents the RKHS. For extending correntropy to metric. Furthermore, the Correntropy can be induced as a metric named textbf{Correntropy Induced Metric (CIM)}, which can be expressed as follows:
\begin{align}
&\text{CIM}(p,q)=(V_\sigma(0)-V_\sigma(p,q))^\frac{1}{2}. \label{CIM}
\end{align}

\section{Our Approach}
This section analyzes PPO-KL's asymmetric surrogate and gradient computation to identify its weaknesses. Then, we propose the PPO-CIM to address these weaknesses.

\subsection{Asymmetry and Instability of PPO-KL}
Compared to TRPO, PPO-KL transforms the constraint condition to a penalty function multiplying an adaptive penalty parameter $\beta$ and adds in the surrogate function. However, the KL divergence is asymmetric and does not obey triangle inequality~\citep{KL2}. In the case of normal distribution, the asymmetry of KL divergence will decrease the learning efficiency. To illustrate it, we give the asymmetric difference of KL-divergences in the following lemma:
\begin{lemma}\label{lemma2}
For any multi-dimensional normal distributions $\boldsymbol{\pi}_{\boldsymbol{\theta}_i}(\boldsymbol{a}|\boldsymbol{s})\sim\mathcal{N}(a|\mathbf{F}_{\boldsymbol{\theta}_i}^{\boldsymbol{\mu}},\mathbf{F}_{\theta_i}^{\boldsymbol{\Sigma}}),\boldsymbol{\pi}_{\boldsymbol{\theta}_j}(\boldsymbol{a}|\boldsymbol{s})\sim\mathcal{N}(a|\mathbf{F}_{\theta_j}^{\boldsymbol{\mu}},\mathbf{F}_{\theta_j}^{\boldsymbol{\Sigma}})$. Denote the asymmetry of the KL divergence is $\Delta D_{\text{KL}}(\boldsymbol{\pi}_{\boldsymbol{\theta}_i},\boldsymbol{\pi}_{\boldsymbol{\theta}_j}):=\kl{\boldsymbol{\pi}_{\boldsymbol{\theta}_i}(\cdot|\boldsymbol{s})}{\boldsymbol{\pi}_{\boldsymbol{\theta}_j}(\cdot|\boldsymbol{s})}-\kl{\boldsymbol{\pi}_{\boldsymbol{\theta}_j}(\cdot|\boldsymbol{s})}{\boldsymbol{\pi}_{\boldsymbol{\theta}_i}(\cdot|\boldsymbol{s})}$. Then the asymmetry can be represented as follows:
\begin{align}
&\Delta D_{\rm{KL}}(\boldsymbol{\pi}_{\boldsymbol{\theta}_i},\boldsymbol{\pi}_{\boldsymbol{\theta}_j})\nonumber \\
=&\log\frac{|\mathbf{F}_{\theta_i}^{\boldsymbol{\Sigma}}|}{|\mathbf{F}_{\theta_j}^{\boldsymbol{\Sigma}}|}+\frac{1}{2}\big[\rm{Trace}\big((\mathbf{F}_{\theta_i}^{\boldsymbol{\Sigma}})^{-1}\mathbf{F}_{\theta_j}^{\boldsymbol{\Sigma}}+(\mathbf{F}_{\theta_j}^{\boldsymbol{\Sigma}})^{-1}(\mathbf{F}_{\theta_j}^{\boldsymbol{\mu}}-\mathbf{F}_{\theta_i}^{\boldsymbol{\mu}})\nonumber \\
&\cdot(\mathbf{F}_{\theta_j}^{\boldsymbol{\mu}}-\mathbf{F}_{\theta_i}^{\boldsymbol{\mu}})^\top\big)-\rm{Trace}\big((\mathbf{F}_{\theta_j}^{\boldsymbol{\Sigma}})^{-1}\mathbf{F}_{\theta_i}^{\boldsymbol{\Sigma}}+(\mathbf{F}_{\theta_i}^{\boldsymbol{\Sigma}})^{-1}(\mathbf{F}_{\theta_i}^{\boldsymbol{\mu}}-\mathbf{F}_{\theta_j}^{\boldsymbol{\mu}})\nonumber \\
&\cdot(\mathbf{F}_{\theta_i}^{\boldsymbol{\mu}}-\mathbf{F}_{\theta_j}^{\boldsymbol{\mu}})^\top\big)\big]. \nonumber
\end{align}
\end{lemma}
The proof of Lemma \ref{lemma2} is shown in Appendix~\ref{app2}. Then, we show whether the asymmetric KL-divergence will affect the learning of PPO-KL in the following theorem: 
\begin{theorem}\label{the1}
In the PPO-KL algorithm, consider the adaptive parameter $\beta$ as a constant, for any two $n$ dimensional policies $\boldsymbol{\pi}_{\boldsymbol{\theta}_i}(\boldsymbol{a}|\boldsymbol{s})\sim\mathcal{N}(a|\mathbf{F}_{\theta_i}^{\boldsymbol{\mu}},\mathbf{F}_{\theta_i}^{\boldsymbol{\Sigma}}),\boldsymbol{\pi}_{\boldsymbol{\theta}_j}(\boldsymbol{a}|\boldsymbol{s})\sim\mathcal{N}(a|\mathbf{F}_{\theta_j}^{\boldsymbol{\mu}},\mathbf{F}_{\theta_j}^{\boldsymbol{\Sigma}})$. If the policy  $\boldsymbol{\pi}_{\boldsymbol{\theta}_j}$ is better than $\boldsymbol{\pi}_{\boldsymbol{\theta}_i}$ and following inequalities are hold:
\begin{align}
&\beta \sum_{k=1}^n \bigg(\|\log h_k\| + \frac{1-h_k^2}{2h_k}\bigg)\nonumber \\
>&\Eb{\boldsymbol{\pi}_{\boldsymbol{\theta}_i}}{\frac{\boldsymbol{\pi}_{\boldsymbol{\theta}_j}(\boldsymbol{a}|\boldsymbol{s})}{\boldsymbol{\pi}_{\boldsymbol{\theta}_i}(\boldsymbol{a}|\boldsymbol{s})}\Api(s,a)}-\Eb{\boldsymbol{\pi}_{\boldsymbol{\theta}_j}}{\frac{\boldsymbol{\pi}_{\boldsymbol{\theta}_i}(\boldsymbol{a}|\boldsymbol{s})}{\boldsymbol{\pi}_{\boldsymbol{\theta}_j}(\boldsymbol{a}|\boldsymbol{s})}\Api(s,a)},\label{eqthe1}\\
&\Eb{\boldsymbol{\pi}_{\boldsymbol{\theta}_i}}{\frac{\boldsymbol{\pi}_{\boldsymbol{\theta}_j}(\boldsymbol{a}|\boldsymbol{s})}{\boldsymbol{\pi}_{\boldsymbol{\theta}_i}(\boldsymbol{a}|\boldsymbol{s})}{A}^{\boldsymbol{\pi}}(\boldsymbol{s},\boldsymbol{a})}>\Eb{\boldsymbol{\pi}_{\boldsymbol{\theta}_j}}{\frac{\boldsymbol{\pi}_{\boldsymbol{\theta}_i}(\boldsymbol{a}|\boldsymbol{s})}{\boldsymbol{\pi}_{\boldsymbol{\theta}_j}(\boldsymbol{a}|\boldsymbol{s})}{A}^{\boldsymbol{\pi}}(\boldsymbol{s},\boldsymbol{a})}.\nonumber
\end{align}
Then, the surrogate functions will show the contrary result:
\begin{align}
&\mathcal{L}^{\rm KL}(\boldsymbol{\pi}_{\boldsymbol{\theta}_i},\boldsymbol{\pi}_{\boldsymbol{\theta}_j})>\mathcal{L}^{\rm KL}(\boldsymbol{\pi}_{\boldsymbol{\theta}_j},\boldsymbol{\pi}_{\boldsymbol{\theta}_i}). \label{L1L2}
\end{align}
where $h_k=\frac{\mathbf{F}_{\theta_i,kk}^{\boldsymbol{\Sigma}}}{\mathbf{F}_{\theta_j,kk}^{\boldsymbol{\Sigma}}}$ represents the ratio of the $k$th diagonal components of the variances of policies.
\end{theorem}
\begin{proof}
By using the result of lemma~\ref{lemma2}, the difference between the two policies $\boldsymbol{\pi}_{\boldsymbol{\theta}_i},\boldsymbol{\pi}_{\boldsymbol{\theta}_j}$ can be represented as follows:
\begin{align}
&\Delta D_{\rm KL}(\boldsymbol{\pi}_{\boldsymbol{\theta}_i},\boldsymbol{\pi}_{\boldsymbol{\theta}_j})\nonumber \\
=&\kl{\boldsymbol{\pi}_{\boldsymbol{\theta}_i}(\cdot|\boldsymbol{s})}{\boldsymbol{\pi}_{\boldsymbol{\theta}_j}(\cdot|\boldsymbol{s})}-\kl{\boldsymbol{\pi}_{\boldsymbol{\theta}_j}(\cdot|\boldsymbol{s})}{\boldsymbol{\pi}_{\boldsymbol{\theta}_i}(\cdot|\boldsymbol{s})}\nonumber \\
=&\log\frac{|\mathbf{F}_{\theta_i}^{\boldsymbol{\Sigma}}|}{|\mathbf{F}_{\theta_j}^{\boldsymbol{\Sigma}}|}+\frac{1}{2}(\text{Trace}[(\mathbf{F}_{\theta_i}^{\boldsymbol{\Sigma}})^{-1}\mathbf{F}_{\theta_j}^{\boldsymbol{\Sigma}}+(\mathbf{F}_{\theta_j}^{\boldsymbol{\Sigma}})^{-1}(\mathbf{F}_{\theta_j}^{\boldsymbol{\mu}}-\mathbf{F}_{\theta_i}^{\boldsymbol{\mu}})\nonumber\\
&\cdot(\mathbf{F}_{\theta_j}^{\boldsymbol{\mu}}-\mathbf{F}_{\theta_i}^{\boldsymbol{\mu}})^{\top}]-\text{Trace}[(\mathbf{F}_{\theta_j}^{\boldsymbol{\Sigma}})^{-1}\mathbf{F}_{\theta_i}^{\boldsymbol{\Sigma}}+(\mathbf{F}_{\theta_i}^{\boldsymbol{\Sigma}})^{-1}\nonumber \\
&\cdot(\mathbf{F}_{\theta_i}^{\boldsymbol{\mu}}-\mathbf{F}_{\theta_j}^{\boldsymbol{\mu}})(\mathbf{F}_{\theta_i}^{\boldsymbol{\mu}}-\mathbf{F}_{\theta_j}^{\boldsymbol{\mu}})^{\top}]),\label{eq12} 
\end{align}
By using the additivity and commutativity of the trace operator, the Eq. (\ref{eq12}) can be rewritten as follows:
\begin{align}
&\Delta D_{\rm KL}(\boldsymbol{\pi}_{\boldsymbol{\theta}_i},\boldsymbol{\pi}_{\boldsymbol{\theta}_j})\nonumber \\
=&\log\frac{|\mathbf{F}_{\theta_i}^{\boldsymbol{\Sigma}}|}{|\mathbf{F}_{\theta_j}^{\boldsymbol{\Sigma}}|}+\frac{1}{2}(\text{Trace}[(\mathbf{F}_{\theta_i}^{\boldsymbol{\Sigma}})^{-1}\mathbf{F}_{\theta_j}^{\boldsymbol{\Sigma}}]-\text{Trace}[(\mathbf{F}_{\theta_j}^{\boldsymbol{\Sigma}})^{-1}\mathbf{F}_{\theta_i}^{\boldsymbol{\Sigma}}]\nonumber \\
&+\text{Trace}[(\mathbf{F}_{\theta_j}^{\boldsymbol{\mu}}-\mathbf{F}_{\theta_i}^{\boldsymbol{\mu}})^{\top}(\mathbf{F}_{\theta_j}^{\boldsymbol{\Sigma}})^{-1}(\mathbf{F}_{\theta_j}^{\boldsymbol{\mu}}-\mathbf{F}_{\theta_i}^{\boldsymbol{\mu}})]\nonumber\\
&-\text{Trace}[(\mathbf{F}_{\theta_j}^{\boldsymbol{\mu}}-\mathbf{F}_{\theta_i}^{\boldsymbol{\mu}})^{\top}(\mathbf{F}_{\theta_i}^{\boldsymbol{\Sigma}})^{-1}(\mathbf{F}_{\theta_j}^{\boldsymbol{\mu}}-\mathbf{F}_{\theta_i}^{\boldsymbol{\mu}})])\nonumber\\
=&\log\frac{|\mathbf{F}_{\theta_i}^{\boldsymbol{\Sigma}}|}{|\mathbf{F}_{\theta_j}^{\boldsymbol{\Sigma}}|}+\frac{1}{2}(\text{Trace}[(\mathbf{F}_{\theta_i}^{\boldsymbol{\Sigma}})^{-1}\mathbf{F}_{\theta_j}^{\boldsymbol{\Sigma}}]-\text{Trace}\big[\big(\mathbf{F}_{\theta_j}^{\boldsymbol{\Sigma}}\big)^{-1}\mathbf{F}_{\theta_i}^{\boldsymbol{\Sigma}}\big]\nonumber\\
&-\text{Trace}[(\mathbf{F}_{\theta_j}^{\boldsymbol{\mu}}-\mathbf{F}_{\theta_i}^{\boldsymbol{\mu}})^{\top}((\mathbf{F}_{\theta_i}^{\boldsymbol{\Sigma}})^{-1}-(\mathbf{F}_{\theta_j}^{\boldsymbol{\Sigma}})^{-1})(\mathbf{F}_{\theta_j}^{\boldsymbol{\mu}}-\mathbf{F}_{\theta_i}^{\boldsymbol{\mu}})]\nonumber\\
=&\log\frac{|\mathbf{F}_{\theta_i}^{\boldsymbol{\Sigma}}|}{|\mathbf{F}_{\theta_j}^{\boldsymbol{\Sigma}}|}+\frac{1}{2}\bigg(\sum_{k=1}^n\frac{\mathbf{F}_{\theta_j,kk}^{\boldsymbol{\Sigma}}}{\mathbf{F}_{\theta_i,kk}^{\boldsymbol{\Sigma}}}-\sum_{k=1}^n\frac{\mathbf{F}_{\theta_i,kk}^{\boldsymbol{\Sigma}}}{\mathbf{F}_{\theta_j,kk}^{\boldsymbol{\Sigma}}}\nonumber\\
&-\sum_{k=1}^n((\mathbf{F}_{\theta_i,kk}^{\boldsymbol{\Sigma}})^{-1}-(\mathbf{F}_{\theta_j,kk}^{\boldsymbol{\Sigma}})^{-1})(\mathbf{F}_{\theta_j,k}^{\boldsymbol{\mu}}-\mathbf{F}_{\theta_i,k}^{\boldsymbol{\mu}})^2\bigg)\label{eq13}\\
\geq&\log\frac{|\mathbf{F}_{\theta_i}^{\boldsymbol{\Sigma}}|}{|\mathbf{F}_{\theta_j}^{\boldsymbol{\Sigma}}|}+\frac{1}{2}\bigg(\sum_{k=1}^n\frac{\mathbf{F}_{\theta_j,kk}^{\boldsymbol{\Sigma}}}{\mathbf{F}_{\theta_i,kk}^{\boldsymbol{\Sigma}}}-\sum_{k=1}^n\frac{\mathbf{F}_{\theta_i,kk}^{\boldsymbol{\Sigma}}}{\mathbf{F}_{\theta_j,kk}^{\boldsymbol{\Sigma}}}\nonumber\\
&-\sum_{k=1}^n\|(\mathbf{F}_{\theta_i,kk}^{\boldsymbol{\Sigma}})^{-1}-(\mathbf{F}_{\theta_j,kk}^{\boldsymbol{\Sigma}})^{-1}\|(\mathbf{F}_{\theta_j,k}^{\boldsymbol{\mu}}-\mathbf{F}_{\theta_i,k}^{\boldsymbol{\mu}})^2\bigg)\label{eq14}
\end{align}
Because the function $\log$ is Lipschitz continuous, for any $(\mathbf{F}_{\theta_i,kk}^{\boldsymbol{\Sigma}})^{-1},(\mathbf{F}_{\theta_j,kk}^{\boldsymbol{\Sigma}})^{-1}\in\mathbb{R}^+$, exists a positive number $L\in\mathbb{R}^+$, such that the following inequality holds:
\begin{align}
&\|\log (\mathbf{F}_{\theta_i,kk}^{\boldsymbol{\Sigma}})^{-1}-\log (\mathbf{F}_{\theta_j,kk}^{\boldsymbol{\Sigma}})^{-1}\|\nonumber\\
\leq& L\|(\mathbf{F}_{\theta_i,kk}^{\boldsymbol{\Sigma}})^{-1}
-(\mathbf{F}_{\theta_j,kk}^{\boldsymbol{\Sigma}})^{-1}\|\label{eq15}
\end{align}
Therefore, combining Eq.(\ref{eq14}) and (\ref{eq15}), the following inequality is holds:
\begin{align}
&\Delta D_{\rm KL}(\boldsymbol{\pi}_{\boldsymbol{\theta}_i},\boldsymbol{\pi}_{\boldsymbol{\theta}_j})\nonumber \\
\geq&\log\frac{|\mathbf{F}_{\theta_i}^{\boldsymbol{\Sigma}}|}{|\mathbf{F}_{\theta_j}^{\boldsymbol{\Sigma}}|}+\frac{1}{2}\bigg(\sum_{k=1}^n\frac{\mathbf{F}_{\theta_j,kk}^{\boldsymbol{\Sigma}}}{\mathbf{F}_{\theta_i,kk}^{\boldsymbol{\Sigma}}}-\sum_{k=1}^n\frac{\mathbf{F}_{\theta_i,kk}^{\boldsymbol{\Sigma}}}{\mathbf{F}_{\theta_j,kk}^{\boldsymbol{\Sigma}}}\nonumber\\
&-\sum_{k=1}^n\frac{1}{L}\|\log (\mathbf{F}_{\theta_i,kk}^{\boldsymbol{\Sigma}})^{-1}-\log (\mathbf{F}_{\theta_j,kk}^{\boldsymbol{\Sigma}})^{-1}\|(\mathbf{F}_{\theta_j,k}^{\boldsymbol{\mu}}-\mathbf{F}_{\theta_i,k}^{\boldsymbol{\mu}})^2\bigg)\nonumber\\
\geq&\log\frac{|\mathbf{F}_{\theta_i}^{\boldsymbol{\Sigma}}|}{|\mathbf{F}_{\theta_j}^{\boldsymbol{\Sigma}}|}+\frac{1}{2}\bigg(\sum_{k=1}^n\frac{\mathbf{F}_{\theta_j,kk}^{\boldsymbol{\Sigma}}}{\mathbf{F}_{\theta_i,kk}^{\boldsymbol{\Sigma}}}-\sum_{k=1}^n\frac{\mathbf{F}_{\theta_i,kk}^{\boldsymbol{\Sigma}}}{\mathbf{F}_{\theta_j,kk}^{\boldsymbol{\Sigma}}}\nonumber\\
&-\sum_{k=1}^n\frac{1}{L}\bigg\|\log \frac{\mathbf{F}_{\theta_j,kk}^{\boldsymbol{\Sigma}}}{\mathbf{F}_{\theta_i,kk}^{\boldsymbol{\Sigma}}}\bigg\|(\mathbf{F}_{\theta_j,k}^{\boldsymbol{\mu}}-\mathbf{F}_{\theta_i,k}^{\boldsymbol{\mu}})^2\bigg)\nonumber\\
\geq&\log\frac{|\mathbf{F}_{\theta_i}^{\boldsymbol{\Sigma}}|}{|\mathbf{F}_{\theta_j}^{\boldsymbol{\Sigma}}|}+\frac{1}{2}\bigg(\sum_{k=1}^n\frac{\mathbf{F}_{\theta_j,kk}^{\boldsymbol{\Sigma}}}{\mathbf{F}_{\theta_i,kk}^{\boldsymbol{\Sigma}}}-\sum_{k=1}^n\frac{\mathbf{F}_{\theta_i,kk}^{\boldsymbol{\Sigma}}}{\mathbf{F}_{\theta_j,kk}^{\boldsymbol{\Sigma}}}\nonumber\\
&-\bigg[\sum_{k=1}^n\frac{1}{L}\bigg\|\log \frac{\mathbf{F}_{\theta_j,kk}^{\boldsymbol{\Sigma}}}{\mathbf{F}_{\theta_i,kk}^{\boldsymbol{\Sigma}}}\bigg\|\bigg]\bigg[\sum_{k=1}^n(\mathbf{F}_{\theta_j,k}^{\boldsymbol{\mu}}-\mathbf{F}_{\theta_i,k}^{\boldsymbol{\mu}})^2\bigg]\bigg)\label{eq16}
\end{align}

Denote $h_k=\frac{\mathbf{F}_{\theta_i,kk}^{\boldsymbol{\Sigma}}}{\mathbf{F}_{\theta_j,kk}^{\boldsymbol{\Sigma}}}$ and $\|\Delta \mathbf{F}_{\theta}^{\boldsymbol{\mu}}\|^2=(\mathbf{F}_{\theta_i}^{\boldsymbol{\mu}}-\mathbf{F}_{\theta_j}^{\boldsymbol{\mu}})^2=\sum_{k=1}^n(\mathbf{F}_{\theta_j,k}^{\boldsymbol{\mu}}-\mathbf{F}_{\theta_i,k}^{\boldsymbol{\mu}})^2$, then Eq. (\ref{eq16}) can be represented as follows:
\begin{align}
&\Delta D_{\rm KL}(\boldsymbol{\pi}_{\boldsymbol{\theta}_i},\boldsymbol{\pi}_{\boldsymbol{\theta}_j})\nonumber \\
\geq&\sum_{k=1}^n \log h_k + \frac{1}{2}\big(\sum_{k=1}^n h_k ^{-1}-\sum_{k=1}^n h_k - \sum_{k=1}^n \frac{1}{L}\|-\log h_k\|\nonumber\\
&\cdot\|\Delta \mathbf{F}_{\theta}^{\boldsymbol{\mu}}\|^2\big) \nonumber\\
\geq&\sum_{k=1}^n \bigg(\frac{2L-\|\Delta \mathbf{F}_{\theta}^{\boldsymbol{\mu}}\|^2}{2L}\|\log h_k\| + \frac{1-h_k^2}{2h_k}\bigg)
\end{align}

The Lipschtiz continuous constant $L$ can be any positive number if and only if inequality (\ref{eq15}) is held. Therefore, exists a larger enough number $L'\in\mathbb{R}^+$, such that: for any $L>L'$, the equality $\frac{2L-\|\Delta \mathbf{F}_{\theta}^{\boldsymbol{\mu}}\|^2}{2L}=1$ always holds. So, the above inequality can be represented as follows:
\begin{align}
&\Delta D_{\rm KL}(\boldsymbol{\pi}_{\boldsymbol{\theta}_i},\boldsymbol{\pi}_{\boldsymbol{\theta}_j})\nonumber \\
\geq&\sum_{k=1}^n \log h_k + \frac{1}{2}\big(\sum_{k=1}^n h_k ^{-1}-\sum_{k=1}^n h_k - \sum_{k=1}^n \frac{1}{L}\|-\log h_k\|\nonumber\\
&\cdot\|\Delta \mathbf{F}_{\theta}^{\boldsymbol{\mu}}\|^2\big) \nonumber\\
\geq&\sum_{k=1}^n \bigg(\|\log h_k\| + \frac{1-h_k^2}{2h_k}\bigg)
\end{align}

Therefore, when the condition (\ref{eqthe1}) is satisfies, we have following inequalities hold:
\begin{align}
&\beta \kl{\boldsymbol{\pi}_{\boldsymbol{\theta}_i}}{\boldsymbol{\pi}_{\boldsymbol{\theta}_j}}-\beta \kl{\boldsymbol{\pi}_{\boldsymbol{\theta}_j}}{\boldsymbol{\pi}_{\boldsymbol{\theta}_i}}\nonumber \\
>&\beta \sum_{k=1}^n \bigg(\|\log h_k\| + \frac{1-h_k^2}{2h_k}\bigg)\nonumber\\
>&\Eb{\boldsymbol{\pi}_{\boldsymbol{\theta}_i}}{\frac{\boldsymbol{\pi}_{\boldsymbol{\theta}_j}(\boldsymbol{a}|\boldsymbol{s})}{\boldsymbol{\pi}_{\boldsymbol{\theta}_i}(\boldsymbol{a}|\boldsymbol{s})}\Api(\boldsymbol{s},\boldsymbol{a})}-\Eb{\boldsymbol{\pi}_{\boldsymbol{\theta}_j}}{\frac{\boldsymbol{\pi}_{\boldsymbol{\theta}_i}(\boldsymbol{a}|\boldsymbol{s})}{\boldsymbol{\pi}_{\boldsymbol{\theta}_j}(\boldsymbol{a}|\boldsymbol{s})}\Api(\boldsymbol{s},\boldsymbol{a})}. \label{eq:ieq1} 
\end{align}

Next, transpose Eq.~(\ref{eq:ieq1}), we have:
\begin{align}
&\mathcal{L}^{\text{KL}}(\boldsymbol{\pi}_{\boldsymbol{\theta}_i},\boldsymbol{\pi}_{\boldsymbol{\theta}_j})\nonumber \\
=&\Eb{\boldsymbol{\pi}_{\boldsymbol{\theta}_j}}{\frac{\boldsymbol{\pi}_{\boldsymbol{\theta}_i}(\boldsymbol{a}|\boldsymbol{s})}{\boldsymbol{\pi}_{\boldsymbol{\theta}_j}(\boldsymbol{a}|\boldsymbol{s})}\Api(\boldsymbol{s},\boldsymbol{a})}-\beta \kl{\boldsymbol{\pi}_{\boldsymbol{\theta}_j}}{\boldsymbol{\pi}_{\boldsymbol{\theta}_i}}\nonumber \\
>&\Eb{\boldsymbol{\pi}_{\boldsymbol{\theta}_i}}{\frac{\boldsymbol{\pi}_{\boldsymbol{\theta}_j}(\boldsymbol{a}|\boldsymbol{s})}{\boldsymbol{\pi}_{\boldsymbol{\theta}_i}(\boldsymbol{a}|\boldsymbol{s})}\Api(\boldsymbol{s},\boldsymbol{a})}-\beta \kl{\boldsymbol{\pi}_{\boldsymbol{\theta}_i}}{\boldsymbol{\pi}_{\boldsymbol{\theta}_j}}\nonumber\\
>&\mathcal{L}^{\text{KL}}(\boldsymbol{\pi}_{\boldsymbol{\theta}_j},\boldsymbol{\pi}_{\boldsymbol{\theta}_i}).
\end{align}

It illustrates that the policy $\boldsymbol{\pi}_{\boldsymbol{\theta}_i}$ better than $\boldsymbol{\pi}_{\boldsymbol{\theta}_j}$ in surrogate functions: 
\begin{align}
&\mathcal{L}^{\text{KL}}(\boldsymbol{\pi}_{\boldsymbol{\theta}_i},\boldsymbol{\pi}_{\boldsymbol{\theta}_j})>\mathcal{L}^{\text{KL}}(\boldsymbol{\pi}_{\boldsymbol{\theta}_j},\boldsymbol{\pi}_{\boldsymbol{\theta}_i}). \label{L1L2}
\end{align}
\end{proof}
Theorem~\ref{the1} reveals that when the asymmetry of KL divergence is significant enough, we can't guarantee policy improvement by improving PPO-KL's surrogate. Because the asymmetry of KL divergence is more significant than the estimated value function, this asymmetry will misguide the policy updated and introduce unstable learning because the agent can not constantly improve their performance through the PPO-KL. In high dimensional continuous control tasks, with the increase of the action number $n$, the asymmetric part will be increased, making the situation easy to happen. In addition, the computational complexity of KL-divergence is high, which reduces the learning efficiency. In the next part, from the perspective of inner product space, we will replace KL-divergence with the symmetric metric (CIM) of probability distribution to compensate for the above problems and propose the PPO-CIM.

\subsection{The propose of PPO-CIM}
In this section, we expand the surrogate of PPO-KL as an inner product format and induce the PPO-CIM. Firstly, we analyze the KL-divergence constraint of the PPO-KL's surrogate. By using Taylor's theorem, the KL-divergence can be approximated as follows~\citep{trpo}:
\begin{align}
D_{\text{KL}}(\boldsymbol{\pi}_{\boldsymbol{\theta}}(\cdot|\boldsymbol{s})\|\boldsymbol{\pi}_{ \boldsymbol{\theta}_{\text{old}}}(\cdot|\boldsymbol{s}))\approx\half (\boldsymbol{\theta} - \boldsymbol{\theta}_{\text{old}})^{\top} \mathbf{F}(\boldsymbol{\theta}) (\boldsymbol{\theta} - \boldsymbol{\theta}_{\text{old}})\label{kla}
\end{align}
where $\mathbf{F}(\theta)$ is the Fish Information Matrix (FIM), which can be represented as follows:
\begin{align}
\{\mathbf{F}(\boldsymbol{\theta})\}_{ij}= \frac{\partial}{\partial\boldsymbol{\theta}_{i}}\frac{\partial}{\partial\boldsymbol{\theta}_{j}} \Eb{\boldsymbol{s} \sim \rhopi} { \kl{\boldsymbol{\pi}_{\boldsymbol{\theta}}}{\boldsymbol{\pi}_{ \boldsymbol{\theta}_{\text{old}}}} }.\nonumber
\end{align}

By using the approximation (\ref{kla}), the PPO-KL can be approximated as:
\begin{align}
&\mathcal{L}^{\rm KL}\nonumber\\
\approx&\mathbb{E}_{\boldsymbol{\pi}_{\boldsymbol{\theta}_{\rm old}}}\bigg[\frac{\boldsymbol{\pi}_{\boldsymbol{\theta}}}{\boldsymbol{\pi}_{\boldsymbol{\theta}_{\rm old}}}\hat{A}^{\boldsymbol{\pi}}(s,a)-\beta\text{D}_{\text{KL}}(\boldsymbol{\pi}_{ \boldsymbol{\theta}_{\text{old}}}\|\boldsymbol{\pi})\bigg]\nonumber\\
\approx& \mathbb{E}_{\boldsymbol{\pi}_{\boldsymbol{\theta}_{\rm old}}}\bigg[\frac{\boldsymbol{\pi}_{\boldsymbol{\theta}}}{\boldsymbol{\pi}_{\boldsymbol{\theta}_{\rm old}}}\hat{A}^{\boldsymbol{\pi}}(s,a)-\beta\frac{1}{2}(\boldsymbol{\theta}-\boldsymbol{\theta}_{\text{old}})^\top \textbf{F}(\boldsymbol{\theta})(\boldsymbol{\theta}-\boldsymbol{\theta}_{\text{old}})\bigg]
\end{align}
By applying the matrix decomposition on the FIM $\textbf{F}(\boldsymbol{\theta})$, the approximation of the KL-divergence can be decomposed as follows:
\begin{align}
&(\boldsymbol{\theta}-\boldsymbol{\theta}_{\text{old}})^\top \textbf{F}(\boldsymbol{\theta})(\boldsymbol{\theta}-\boldsymbol{\theta}_{\text{old}})\nonumber\\
=& [(\boldsymbol{\theta}-\boldsymbol{\theta}_{\text{old}})^\top \mathbf{U}^\top \mathbf{\Lambda}^{1/2}][\mathbf{\Lambda}^{1/2}\mathbf{U} (\boldsymbol{\theta}-\boldsymbol{\theta}_{\text{old}})]\nonumber\\
=&\boldsymbol{\theta}^\top(\mathbf{\Lambda}^{1/2}\mathbf{U})^\top(\mathbf{\Lambda}^{1/2}\mathbf{U})\boldsymbol{\theta}-2\boldsymbol{\theta}^\top(\mathbf{\Lambda}^{1/2}\mathbf{U})^\top(\mathbf{\Lambda}^{1/2}\mathbf{U})\boldsymbol{\theta}_{\rm old}\nonumber\\
&+\boldsymbol{\theta}_{\rm old}^\top(\mathbf{\Lambda}^{1/2}\mathbf{U})^\top(\mathbf{\Lambda}^{1/2}\mathbf{U})\boldsymbol{\theta}_{\rm old}\label{apd1}
\end{align}
Denote the function $f\subset\mathcal{H}$ as follows:
\begin{align}
f(\boldsymbol{x}) = \mathbf{\Lambda}^{1/2}\mathbf{U} \boldsymbol{x}
\end{align}
Then the approximation (\ref{apd1}) can be represented as following inner product format:
\begin{align}
&(\boldsymbol{\theta}-\boldsymbol{\theta}_{\text{old}})^\top \textbf{F}(\boldsymbol{\theta})(\boldsymbol{\theta}-\boldsymbol{\theta}_{\text{old}})\nonumber\\
=&f(\boldsymbol{\theta})^\top f(\boldsymbol{\theta})+f(\boldsymbol{\theta_{\rm old}})^\top f(\boldsymbol{\theta}_{\rm old})-2f(\boldsymbol{\theta})^\top f(\boldsymbol{\theta}_{\rm old})\nonumber\\
=&[f(\boldsymbol{\theta})^\top f(\boldsymbol{\theta})-f(\boldsymbol{\theta})^\top f(\boldsymbol{\theta_{\rm old}})]\nonumber\\
&+[f(\boldsymbol{\theta_{\rm old}})^\top f(\boldsymbol{\theta_{\rm old}})-f(\boldsymbol{\theta})^\top f(\boldsymbol{\theta_{\rm old}})]
\end{align}

The format of the above equation is consistent with the square of the CIM~(\ref{CIM}). The difference is the approximation of KL-divergence processing in Euclidean space, but CIM is processing in RKHS space. Therefore, by extending the function $f$ to a kernel mapping $\Phi$, such that $\forall x,y$, the equation $\product{\Phi(x),\Phi(y)}=\kappa(x-y)$ holds, we can extend the PPO-KL to a more general representation, which we named it as PPO-CIM:
\begin{align}
&\mathcal{L}_{\boldsymbol{\pi}_{\boldsymbol{\theta}}}^{\text{CIM}}\nonumber \\
=&\Eb{\boldsymbol{\pi}_{\boldsymbol{\theta}_{\text{old}}}}{\frac{\boldsymbol{\pi}_{\boldsymbol{\theta}}(a|s)}{\boldsymbol{\pi}_{\boldsymbol{\theta}_{\text{old}}}(a|s)}\Api(s,a)}- \alpha\cdot \text{CIM}^2({\boldsymbol{\pi}_{\boldsymbol{\theta}_{\text{old}}}(\cdot|s)},{\boldsymbol{\pi}_{\boldsymbol{\theta}}(\cdot|s)}) . \nonumber
\end{align}
where $\alpha\in\mathbb{R}^+$ is a hyper-parameter that determines the effect of the constraint. Compared with PPO-KL, because the CIM is a symmetric metric~\citep{corr}, the PPO-CIM can avoid the unstable learning we mentioned in Theorem~\ref{the1}. Besides, in PPO-CIM, the adaptive mechanism is canceled: the $\alpha$ is fixed and determined by the task's reward. Due to the adaptive mechanism may also lead to a large CIM and

From the computational complexity and trust region perspectives, we give the following two conditions to guide the kernel function's selection of CIM:
\begin{enumerate}
\item The computational complexity of kernel function $\Phi(\cdot): \mathbb{R}^n\to\mathbb{R}$ is less than $O(n^3)$.\label{cond1}
\item The kernel function should satisfies following inequality: $\product{\Phi(\boldsymbol{x}_i),\Phi(\boldsymbol{x}_j)}\leq |\mathcal{A}|\|\boldsymbol{x}_i^\top\log \boldsymbol{x}_j, \forall \boldsymbol{x}_i,\boldsymbol{x}_j\in [0,1]^{|\mathcal{A}|}$.\label{cond2}
\end{enumerate}

\setlength{\parindent}{0pt}\textbf{Computational Complexity:} In the following theorem, we will discuss the gradient's computational complexity of the PPO-KL and PPO-CIM and validate the condition~\ref{cond1} can ensure the PPO-CIM with lower computational complexity:
\setlength{\parindent}{2em}
\begin{theorem}
If condition~\ref{cond1} is satisfied, the computation complexity of the PPO-CIM's gradient is less than that of the PPO-KL.
\end{theorem}

\begin{proof}

By using the approximation (\ref{kla}) and treating the FIM as a constant, the policy gradient of the PPO-KL can be expressed as follows:
\begin{align}
&\nabla \mathcal{L}_{\boldsymbol{\pi}}^{\rm KL}\nonumber\\=&\nabla\mathbb{E}_{\boldsymbol{\pi}_{\boldsymbol{\theta}_{\text{old}}}}\big[\frac{\boldsymbol{\pi}_{\boldsymbol{\theta}}(a|s)}{\boldsymbol{\pi}_{\boldsymbol{\theta}_{\text{old}}}(a|s)}\Api(s,a)-\beta{D}_{\rm KL}(\boldsymbol{\pi}_{\boldsymbol{\theta}_{\text{old}}}\|\boldsymbol{\pi}_{\boldsymbol{\theta}})\big]\nonumber\\
\approx &\mathbb{E}_{\boldsymbol{\boldsymbol{\pi}}_{\theta_{\rm old}}}\bigg[\frac{\nabla\boldsymbol{\boldsymbol{\pi}}_\theta}{\boldsymbol{\boldsymbol{\pi}}_{\theta_{\rm old}}}A^{\boldsymbol{\boldsymbol{\pi}}}(\boldsymbol{s},\boldsymbol{a})-\beta\nabla(\frac{1}{2}(\boldsymbol{\theta}-\boldsymbol{\theta}_{\text{old}})^\top \textbf{F}(\theta_{\text{old}})\nonumber\\
&\cdot(\boldsymbol{\theta}-\boldsymbol{\theta}_{\text{old}}))\bigg]\nonumber\\
= & \mathbb{E}_{\boldsymbol{\boldsymbol{\pi}}_{\theta_{\rm old}}}\bigg[\frac{\nabla\boldsymbol{\boldsymbol{\pi}}_\theta}{\boldsymbol{\boldsymbol{\pi}}_{\theta_{\rm old}}}A^{\boldsymbol{\boldsymbol{\pi}}}(\boldsymbol{s},\boldsymbol{a})\bigg]-\beta\mathbb{E}_{\boldsymbol{\pi}_{\theta_{\rm old}}}\big[ (\boldsymbol{\theta}-\boldsymbol{\theta}_{\rm old})^\top\mathbf{F}(\theta)\nabla\pi_{\theta}\big]\
\end{align}
Consider the FIM and $\nabla \boldsymbol{\pi}_\theta$ as constants, the last item of the PPO-KL's gradient multiplies $n\times n\times 2$ times in total, and adds $(n-1)\times n\times 2$ times in total. The computational complexity is $O(n^2)$. Then, considering the computational complexity of FIM as second-order information, the computational complexity of FIM is equal to the corresponding of the Hessian matrix, which is $O(n^3)$. Compared with FIM, as the first-order information, the computational complexity of $\nabla \boldsymbol{\pi}_\theta$ can be ignored. To sum up, the computational complexity of PPO-KL's gradient is $O(n^3)$.

Next, we consider the policy gradient of PPO-CIM, which can be represented as follows:
\begin{align}
&\nabla \mathcal{L}_{\boldsymbol{\pi}}^{\rm CIM}\nonumber\\
=&\nabla\mathbb{E}_{\boldsymbol{\pi}_{\boldsymbol{\theta}_{\text{old}}}}\bigg[\frac{\boldsymbol{\pi}_{\boldsymbol{\theta}}(a|s)}{\boldsymbol{\pi}_{\boldsymbol{\theta}_{\text{old}}}(a|s)}\Api(s,a)-\alpha {\rm CIM}^2(\boldsymbol{\pi}_{\boldsymbol{\theta}_{\text{old}}},\boldsymbol{\pi}_{\boldsymbol{\theta}})\bigg]\nonumber\\
= &\mathbb{E}_{\boldsymbol{\pi}_{\boldsymbol{\theta}_{\text{old}}}}\bigg[\frac{\nabla \boldsymbol{\pi}_{\boldsymbol{\theta}}}{\boldsymbol{\pi}_{\boldsymbol{\theta}_{\text{old}}}}A^{\boldsymbol{\pi}}(\boldsymbol{s},\boldsymbol{a})\bigg]-\alpha\mathbb{E}\big[\nabla\big(\langle\Phi(0),\Phi(0)\rangle_{\mathcal{F}}\nonumber\\
&-\product{\Phi(\boldsymbol{\pi}_{\theta_{\rm old}}),\Phi(\boldsymbol{\pi}_{\theta})}\big)\bigg]\nonumber\\
= & \mathbb{E}_{\boldsymbol{\pi}_{\boldsymbol{\theta}_{\text{old}}}}\bigg[\frac{\nabla\boldsymbol{\pi}_\theta}{\boldsymbol{\pi}_{\boldsymbol{\theta}_{\text{old}}}}A^{\boldsymbol{\pi}}(\boldsymbol{s},\boldsymbol{a})\bigg]-\alpha\mathbb{E}\bigg[ \product{\Phi(\boldsymbol{\pi}_{\theta_{\rm old}}),\nabla\Phi(\boldsymbol{\pi}_{\theta})}\bigg]\
\end{align}
Similarly, if we don't consider the computational complexity of the kernel function, we can obtain the computational complexity of PPO-CIM's gradient is $O(n)$. Then, if the computational complexity of the kernel function $\Phi(\cdot)$ is more significant than $O(n)$, the computational complexity of PPO-CIM's gradient will be consistent with $\Phi(\cdot)$.

In conclusion, the computational complexity of $\Phi(\cdot)$ should be less than $O(n^3)$, compared with the gradient of PPO-KL.
\end{proof}

\setlength{\parindent}{0pt}\textbf{Trust Region Constrain:} Compared with PPO-Clip, the PPO-CIM can ensure the new policy is always within the trust region, as long as the following trust region condition~\ref{cond2} is satisfied:
\setlength{\parindent}{2em}
\begin{theorem}\label{the2}
In the PPO-CIM algorithm, If the kernel function satisfies the condition \ref{cond2}. Then, the CIM can restrict the new policy within the trust region:
\begin{align}
{\rm CIM}^2(\boldsymbol{\pi}_{\boldsymbol{\theta}_i},\boldsymbol{\pi}_{\boldsymbol{\theta}_j})\geq \text{D}_{{\rm KL}}(\boldsymbol{\pi}_{\boldsymbol{\theta}_i} \| \boldsymbol{\pi}_{\boldsymbol{\theta}_j}),\forall \boldsymbol{\pi}_{\boldsymbol{\theta}_i},\boldsymbol{\pi}_{\boldsymbol{\theta}_j}
\end{align}
\begin{proof}
By using the importance sampling, the KL-divergence between $\boldsymbol{\pi}_{\boldsymbol{\theta}_i}$ and $\boldsymbol{\pi}_{\boldsymbol{\theta}_j}$ can be represented as following:
\begin{align}
&D_{\rm KL}(\boldsymbol{\pi}_{\boldsymbol{\theta}_i} \| \boldsymbol{\pi}_{\boldsymbol{\theta}_j}) \nonumber\\
 = &\mathbb{E}_{\pi_{\boldsymbol{\theta}_i}}\bigg[\log \frac{\boldsymbol{\pi}_{\boldsymbol{\theta}_i}}{\boldsymbol{\pi}_{\boldsymbol{\theta}_j}}\bigg]\nonumber\\
=&\mathbb{E}_{a\sim{\rm Unif(\mathcal{A})}}\bigg[\frac{\boldsymbol{\pi}_{\boldsymbol{\theta}_i}}{1/|\mathcal{A}|}\log \frac{\boldsymbol{\pi}_{\boldsymbol{\theta}_i}}{\boldsymbol{\pi}_{\boldsymbol{\theta}_j}}\bigg]\nonumber\\
=&\mathbb{E}_{a\sim{\rm Unif(\mathcal{A})}}\big[{|\mathcal{A}|\pi_{\boldsymbol{\theta}_i}}(\log {\boldsymbol{\pi}_{\boldsymbol{\theta}_i}}-\log{\boldsymbol{\pi}_{\boldsymbol{\theta}_j}})\big]\label{kl1}
\end{align}
where ${\rm Unif}(\mathcal{A})$ represents the uniform distribution of the action space. Due to the $\boldsymbol{\pi}_{\boldsymbol{\theta}_i},\boldsymbol{\pi}_{\boldsymbol{\theta}_j}$ is defined on the probability space, the $\log \boldsymbol{\pi}_{\boldsymbol{\theta}_i}$ and $\log \boldsymbol{\pi}_{\boldsymbol{\theta}_j}$ are non-positive define. So we can find a more strict inequality from (\ref{kl1}):
\begin{align}
&D_{\rm KL}(\boldsymbol{\pi}_{\boldsymbol{\theta}_i} \| \boldsymbol{\pi}_{\boldsymbol{\theta}_j})\nonumber\\
=&\mathbb{E}_{a\sim{\rm Unif(\mathcal{A})}}\big[{|\mathcal{A}|\boldsymbol{\pi}_{\boldsymbol{\theta}_i}}(\log {\boldsymbol{\pi}_{\boldsymbol{\theta}_i}}-\log{\boldsymbol{\pi}_{\boldsymbol{\theta}_j}})\big]\nonumber\\
\leq&\mathbb{E}_{a\sim{\rm Unif(\mathcal{A})}}\big[-|\mathcal{A}|\boldsymbol{\pi}_{\boldsymbol{\theta}_i}\log{\boldsymbol{\pi}_{\boldsymbol{\theta}_j}}\big]\label{kl2}
\end{align}

If condition~\ref{cond2} is satisfied, then the following inequality holds:
\begin{align}
&{\rm CIM}^2(\boldsymbol{\pi}_{\boldsymbol{\theta}_i},\boldsymbol{\pi}_{\boldsymbol{\theta}_j})\nonumber\\
=&\mathbb{E}[\product{\Phi(0),\Phi(0)}]-\mathbb{E}\big[\langle\Phi(\boldsymbol{\pi}_{\boldsymbol{\theta}_i}),\Phi(\boldsymbol{\pi}_{\boldsymbol{\theta}_j})\rangle_{\mathcal{F}}\big]\nonumber\\
\geq& \mathbb{E}[\product{\Phi(0),\Phi(0)}]-\mathbb{E}[|\mathcal{A}|\boldsymbol{\pi}_{\boldsymbol{\theta}_i}^\top\log \boldsymbol{\pi}_{\boldsymbol{\theta}_j}|]\label{cim1}
\end{align}
Combining inequalities (\ref{kl2}) and (\ref{cim1}), the following inequality is holds:
\begin{align}
&{\rm CIM}^2(\boldsymbol{\pi}_{\boldsymbol{\theta}_i},\boldsymbol{\pi}_{\boldsymbol{\theta}_j})\nonumber\\
\geq& \mathbb{E}[\product{\Phi(0),\Phi(0)}]-\mathbb{E}[|\mathcal{A}|\boldsymbol{\pi}_{\boldsymbol{\theta}_i}^\top\log \boldsymbol{\pi}_{\boldsymbol{\theta}_j}|]\nonumber\\
\geq&\mathbb{E}[\product{\Phi(0),\Phi(0)}]+ D_{\rm KL}(\boldsymbol{\pi}_{\boldsymbol{\theta}_i} \| \boldsymbol{\pi}_{\boldsymbol{\theta}_j})
\end{align}

\end{proof}

\end{theorem}

\subsection{Implementation of PPO-CIM}

Directly replacing the KL divergence by the square of the CIM can not reduce the computational cost significantly due to the computational complexity of $\kappa(\boldsymbol{\pi}_{\boldsymbol{\theta}}-\boldsymbol{\pi}_{\boldsymbol{\theta}_{\rm old}})$ is hard to reduce under $O(n^3)$ and differential operation of some reproducing kernels are also very complex~\citep{HLong}. 

To further reduce the computational complexity, we discuss the approximation technique of CIM. First, we assume the Actor and Critic networks in PPO-CIM satisfy the Lipschtiz continuous:
\begin{assumption}\label{ass1}
Assume the Actor and Critic networks satisfy the following L-Lipschitz continuous:
\begin{equation}
\exists  L\in(0, 1], s.t~\|f_{\boldsymbol{\theta}_{i+1}}-f_{\boldsymbol{\theta}_{i}}\|\leq L\|\boldsymbol{\theta}_{i+1}-\boldsymbol{\theta}_{i}\|
\end{equation}
where $f_{\boldsymbol{\theta}_i}$  is the output of the network with parameter $\boldsymbol{\theta}_i$, $L$ is the Lipschitz constant of neural networks, which is proved bounded in~{\rm \citep{lpsNN}}.
    
\end{assumption}
This assumption can be satisfied when some networks (such as CNN) and the weights of networks are normalized~\citep{lpsNN}.  Therefore, when the assumption~\ref{ass1} is held, the policies and their parameters are satisfied following inequality:
\begin{align}
    \|\boldsymbol{\pi}_{\boldsymbol{\theta}_i}-\boldsymbol{\pi}_{\boldsymbol{\theta}_j}\|\leq L\|\boldsymbol{\theta}_{i}-\boldsymbol{\theta}_{j}\|,0<L\leq1\label{ip1}
\end{align}
Then, according to Eq. (\ref{gkernel}), the Gaussian kernel can be rewritten as following monotonically decreasing function:
    \begin{align}
      \kappa(\boldsymbol{\pi}_{\boldsymbol{\theta}_i}-\boldsymbol{\pi}_{\boldsymbol{\theta}_j})=\kappa(\|\boldsymbol{\pi}_{\boldsymbol{\theta}_i}-\boldsymbol{\pi}_{\boldsymbol{\theta}_j}\|),\label{ip2}
    \end{align}
Combining with Eqs. (\ref{ip1}) and (\ref{ip2}), the following inequalities are held as well:
    \begin{align}
         &\kappa(\|\boldsymbol{\pi}_{\boldsymbol{\theta}_i}-\boldsymbol{\pi}_{\boldsymbol{\theta}_j}\|)\geq \kappa(L\cdot\|\boldsymbol{\theta}_{i}-\boldsymbol{\theta}_{j}\|)\geq\kappa(\|\boldsymbol{\theta}_{i}-\boldsymbol{\theta}_{j}\|)
    \end{align}

By using the definition of the CIM (See Eq. (\ref{CIM})), we can derive that the policies' CIM is less than the corresponding parameters:
\begin{align}
    {\rm CIM}(\boldsymbol{\pi}_{\boldsymbol{\theta}_i},\boldsymbol{\pi}_{\boldsymbol{\theta}_j})\leq  {\rm CIM}(\boldsymbol{\theta}_{i},\boldsymbol{\theta}_{j})
\end{align}
Combining with theorem~\ref{the2}, utilizing the parameters' CIM to replace the policies' CIM can ensure the policy is restricted within the trust region and make the condition~\ref{cond2} easier to reach. After replacing, the surrogate of PPO-CIM can be rewritten as follows:
\begin{align}
&\mathcal{L}_{\boldsymbol{\pi}_{\boldsymbol{\theta}}}^{\text{CIM}}\nonumber \\
\approx&\Eb{\boldsymbol{\pi}_{\boldsymbol{\theta}_{\text{old}}}}{\frac{\boldsymbol{\pi}_{\boldsymbol{\theta}}(a|s)}{\boldsymbol{\pi}_{\boldsymbol{\theta}_{\text{old}}}(a|s)}\Api(s,a)}- \alpha\cdot \text{CIM}^2(\boldsymbol{\theta}_{\text{old}},{\boldsymbol{\theta}}) . \label{CIMe}
\end{align}

Consider the CIM with Gaussian kernel, for any two policies $\boldsymbol{\pi}_{\boldsymbol{\theta}_i},\boldsymbol{\pi}_{\boldsymbol{\theta}_j}$, their CIM square can be represented as follows:
\begin{align}
    {\rm CIM}^{2}({\boldsymbol{\theta}_i},{\boldsymbol{\theta}_j})=\mathbb{E}\Big[\frac{1}{2\pi\sigma} (1-\exp\{-\|\boldsymbol{\theta}_i-\boldsymbol{\theta}_j\|^2/2\sigma^2\})\Big]\label{cim2}
\end{align}

By using the Taylor theorem, the Eq. (\ref{cim2}) can be expanded as follows:
\begin{align}
    {\rm CIM}^{2}({\boldsymbol{\theta}_i},{\boldsymbol{\theta}_j})=\mathbb{E}\Big[\frac{1}{2\pi\sigma}\sum_{n=1}^{+\infty}\frac{\|\boldsymbol{\theta}_i-\boldsymbol{\theta}_j\|^{2n}}{2^n n!}\Big]
\end{align}
To further reduce the computational cost, we select the first and second-order items as the approximations and denote them as $\hat{\rm CIM}_1^{'2},\hat{\rm CIM}_2^{'2}$, respectively:
\begin{align}
    \hat{\rm CIM}_1^{2}(\boldsymbol{\pi}_{\boldsymbol{\theta}_i},\boldsymbol{\pi}_{\boldsymbol{\theta}_j})&\approx \frac{1}{2\pi\sigma}\mathbb{E}\Big[\frac{\|\boldsymbol{\theta}_i-\boldsymbol{\theta}_j\|^2}{2\sigma^2}\Big]\label{CIM1}\\
     \hat{\rm CIM}_2^{2}(\boldsymbol{\pi}_{\boldsymbol{\theta}_i},\boldsymbol{\pi}_{\boldsymbol{\theta}_j})&\approx\frac{1}{2\pi\sigma}\mathbb{E}\bigg[ \frac{\|\boldsymbol{\theta}_i-\boldsymbol{\theta}_j\|^2}{2\sigma^2}-\frac{\|\boldsymbol{\theta}_i-\boldsymbol{\theta}_j\|^4}{8\sigma^4}\bigg]\label{CIM2}
\end{align}

By introducing Eqs. (\ref{CIM1}) and (\ref{CIM2}) into the surrogate (\ref{CIMe}), we can obtain the PPO-CIM's approximation format that we called them as PPO-CIM-1 and PPO-CIM-2, respectively. Take surrogate (\ref{CIMe}) for example; we give the pseudo-code of PPO-CIM in Algorithm~\ref{alg:corr}

\begin{algorithm}
\begin{algorithmic}
\STATE Initial actor and critic networks' parameters: $\boldsymbol{\theta}_0, \phi_0$.\\
\FOR{$i=0,1,2,\dots$ until convergence}
\STATE Running policy $\boldsymbol{\pi}_{\boldsymbol{\theta}_i}$ in the task environment, and storage the trajectories in the set $\mathcal{D}_i=\{\tau_k \}$.
\STATE Optimize follow estimation by mini-batch SGD or Adam:\\
\STATE $\boldsymbol{\theta}_{i+1}$\\$=\mathop{\arg\max}\limits_{\boldsymbol{\theta}} \frac{1}{|D_i|\cdot N} \sum_{\tau\in D_i}\sum_{k=0}^N\frac{\boldsymbol{\pi}_{\boldsymbol{\theta}}}{\boldsymbol{\pi}_{\boldsymbol{\theta}_i}}\hat{A}_t-\alpha \text{CIM}^2( \boldsymbol{\theta}_i, \boldsymbol{\theta})$\\
Update $\phi_k$ through value function $V_\phi$ of Actor-Critic:
\STATE $\phi_{i+1}=\mathop{\arg\min}\limits_{\phi} \frac{1}{|D_i|\cdot N} \sum_{\tau\in D_i}\sum_{k=0}^N(V_{\phi}(s_t)-\hat{R}_t)^2$
\ENDFOR
\end{algorithmic}
\caption{PPO with Correntropy Induced Metric
\label{alg:corr}}
\end{algorithm}

\section{Experiment}
In this section, we visualized the asymmetry of the KL divergence, validated its' negative effects, and compared the proposed algorithms(PPO-CIM, PPO-CIM-1, and PPO-CIM-2) with PPO-KL and PPO-Clip on the continuous-action control tasks that selected from the MuJoCo.
\subsection{Experimental Setting}
\begin{figure*}[htbp]
\centering
\subfloat[]{\includegraphics[width=1.5in]{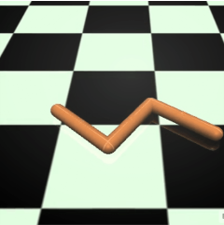}}\;\;
\subfloat[]{\includegraphics[width=1.5in]{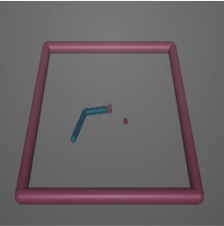}}\;\;\\
\subfloat[]{\includegraphics[width=1.5in]{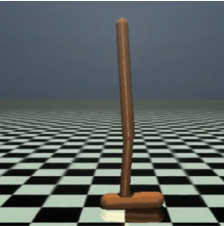}}\;\;
\subfloat[]{\includegraphics[width=1.5in]{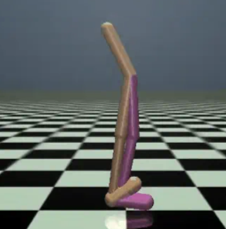}}\;\;
\subfloat[]{\includegraphics[width=1.5in]{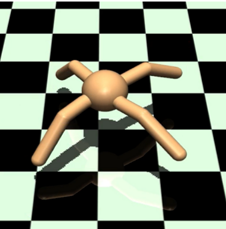}}\;\;
\subfloat[]{\includegraphics[width=1.5in]{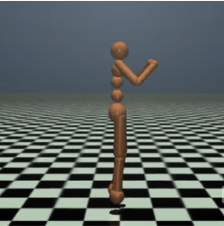}}
\caption{Screenshots of the environments for Mujoco continuous tasks: (a)Swimmer-v2. (b) Reacher-v2. (c) Hopper-v2. (d)Walker2d-v2. (e)Ant-v2. (f)Humanoid-v2}
\label{env}
\end{figure*}

The experiments were conducted on a PC with the following specifications: Windows 11 (64-bit), AMD Ryzen 9 processor, 64GB RAM, and NVIDIA GeForce RTX 3080 GPU, 32GB. To validate the proposed algorithms, we select six continuous-action control tasks as experiment environments from Mujoco (See the screenshots in Fig.~\ref{env}); the details of the tasks are shown in Tab.~\ref{envtab}. The actor and critic networks consisted of two layers of Deep Neural Networks with 128 neurons in each layer. The activation function used was ${\rm tanh}$. The hyperparameters setting of PPO-CIM, PPO-KL, and PPO-Clip can be seen in Tab.~\ref{para}. 
\begin{table*}[t]
\centering
\caption{The information of the four tasks}
\label{envtab}
\begin{tabular}{cccc}
\hline
Task name &Number of actions&Number of states&Target\\ \hline
Swimmer& 2&8& Making a 2D three-link rod swim forward as fast as possible.\\ 
Reacher & 2 & 11& Making a 2D three-link agent to reach a target in
a square-bounded region.\\
Hopper & 11& 3 &Making a 2D one-leg agent hop forward  as fast as possible.\\
Walker2d & 17 &6 &Making a 2D bipedal agent walk forward  as fast as possible.\\
Ant & 111&8&Making a 3D four-legged agent walk forward  as fast as possible.\\
Humanoid& 376&17& Making 3D bipedal agent walk forward  as fast as possible.\\\hline
\end{tabular}
\end{table*}
\begin{table}[ht]
\centering
\caption{The hyperparameters setting of the algorithms}
\label{para}
\begin{tabular}{ccc}
\hline
Algorithms name &Hyperparameters name & Value\\ \hline
PPO-KL& $d_{\text{targ}}$& 0.1\\ 
PPO-KL & $\beta$ & 1\\
PPO-Clip & $\varepsilon$& 0.2 \\
PPO-CIM& $\alpha$ & 2 \\
PPO-KL/Clip/CIM & $\gamma$ & 0.99 \\
PPO-KL/Clip/CIM & Learning rate of Actor & 0.0001 \\
PPO-KL/Clip/CIM & Learning rate of Critic & 0.0001 \\
PPO-KL/Clip/CIM & Update batch size & 256 \\ 
PPO-KL/Clip/CIM & Critic update steps & 10 \\ 
PPO-KL/Clip/CIM & Actor update steps & 10 \\ \hline
\end{tabular}
\end{table}

\subsection{Asymmetry of KL Divergence}
\begin{figure*}[htbp]
\centering
\subfloat[]{\includegraphics[width=1.72in]{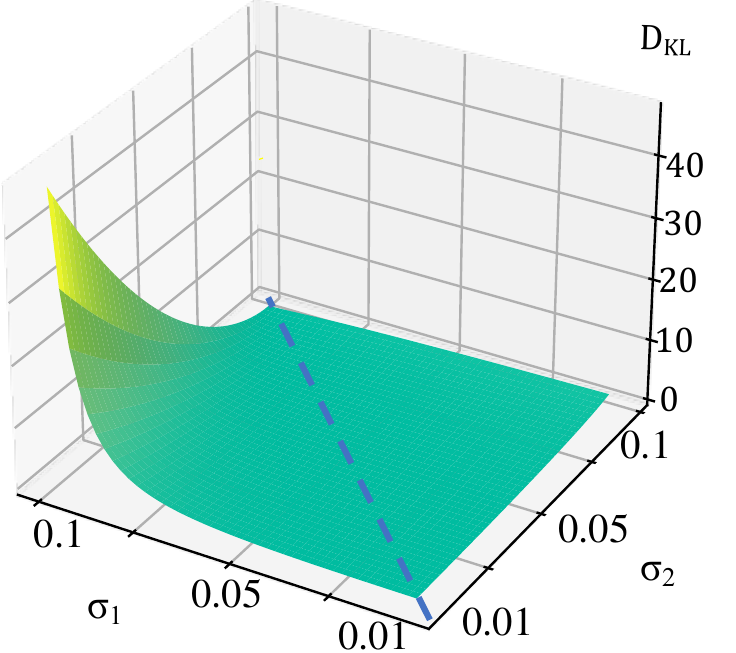}}
\subfloat[]{\includegraphics[width=1.72in]{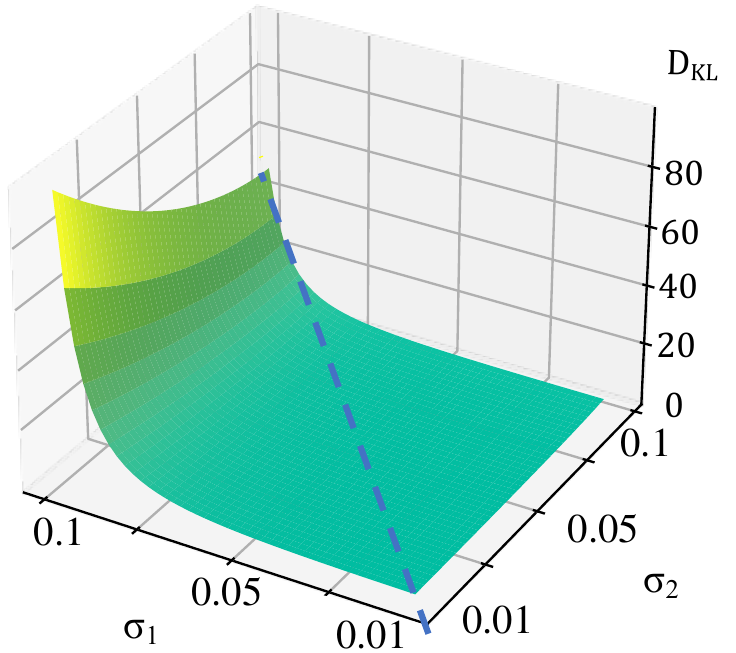}}
\subfloat[]{\includegraphics[width=1.72in]{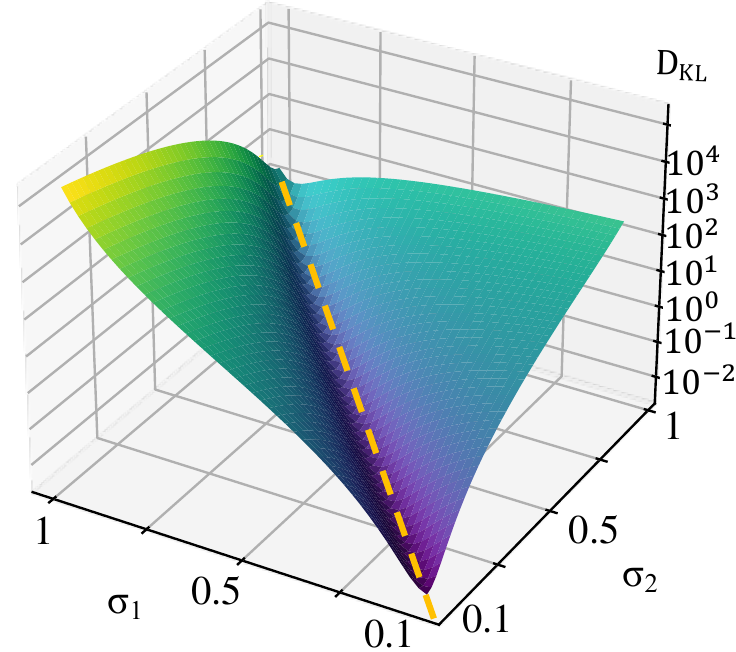}}
\subfloat[]{\includegraphics[width=1.72in]{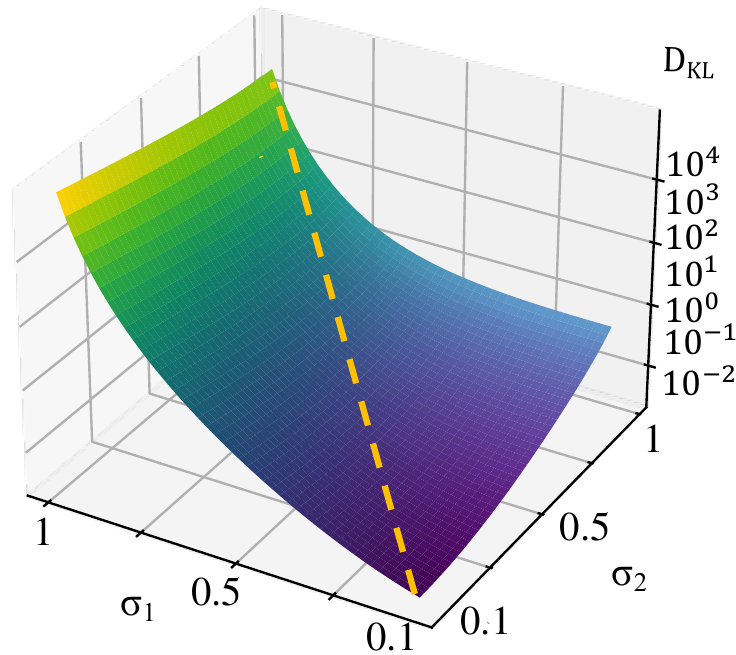}}
\caption{KL-divergence ${\rm D_{KL}}$ between two policies. (a): $\mu_1=\mu_2=1;\sigma\in[0.01,0.1]$; (b): $\mu_1=1,\mu_2=1.1, \sigma\in[0.01,0.1]$; (c): $\mu_1=1,\mu_2=1.1, \sigma\in[0.01,1]$;(d): $\mu_1=1,\mu_2=2, \sigma\in[0.01,1]$.}
\label{KLdiv}
\end{figure*}

To show the changes of the KL-divergence as the policies' parameters difference, consider two 1-D policies $\pi_1\sim\mathcal{N}(\mu_1,\sigma_1),\pi_2\sim\mathcal{N}(\mu_2,\sigma_2)$, then we changes their means and standard variances and visualize their KL divergence as a 3D surface in Fig.~\ref{KLdiv}. 
\begin{figure*}[htbp]
\centering
\subfloat[]{\includegraphics[width=1.72in]{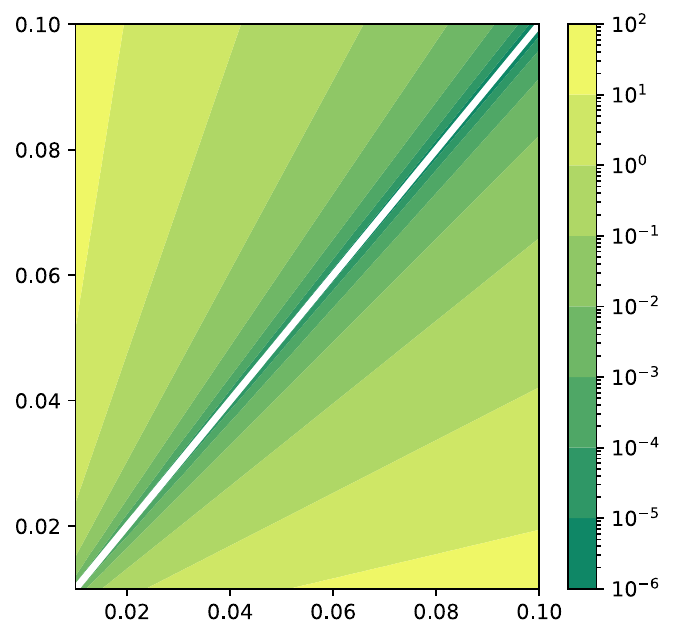}}
\subfloat[]{\includegraphics[width=1.72in]{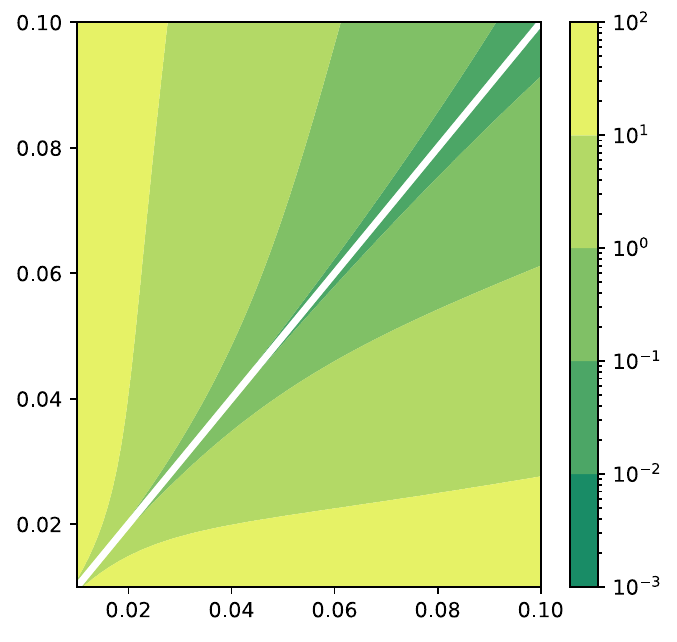}}
\subfloat[]{\includegraphics[width=1.72in]{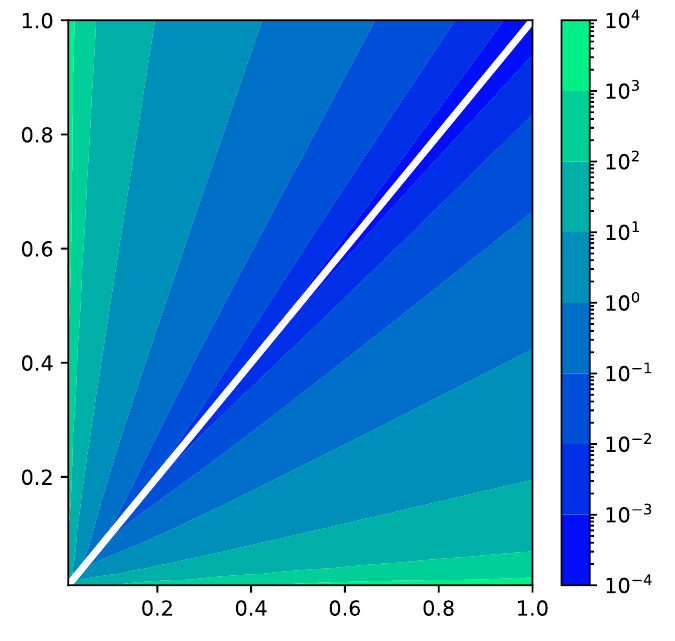}}
\subfloat[]{\includegraphics[width=1.72in]{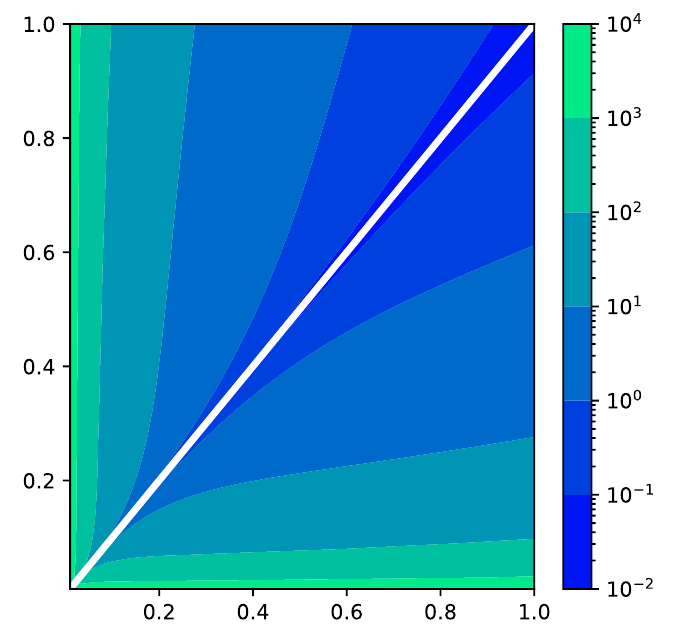}}
\caption{The asymmetry of the KL-divergence $\Delta {\rm D_{KL}}$ between two policies (a): $\mu_1=\mu_2=1, \sigma\in[0.01,0.1]$; (b): $\mu_1=1,\mu_2=1.1, \sigma\in[0.01,0.1]$; (c): $\mu_1=1,\mu_2=1.1, \sigma\in[0.01,1]$; (d): $\mu_1=1,\mu_2=2, \sigma\in[0.01,1]$}
\label{KLdiv2}
\end{figure*}
In the figures, the $x$ and $y$ axis represent the standard variances of $\pi_1,\pi_2$, respectively, the $z$ axis represents the KL-divergence, the dotted lines represent the two policies with the same standard variance, and the point-pair on the surface that symmetry about the line represent the KL-divergence ${\rm D_{KL}}({\pi}_1\|\pi_2)$ and ${\rm D_{KL}}({\pi}_2\|\pi_1)$, respectively. Specifically, Fig.~\ref{KLdiv} (a)-(d) shows that as the differences of means or standard variance increase, the KL divergence rises simultaneously. See the left and right corners in the figures. As the difference between $\sigma_1$ and $\sigma_2$ increases, the KL-divergence increases. We compare Fig.~\ref{KLdiv} (a) and (c) with (b) and (d), respectively. The results illustrate while the $\mu_2$ increased from 1 to 2, the maximum of the KL-divergence increases rapidly, which indicates that the maximum of KL-divergence is positively related to the difference of means. Then, we compare Fig.~\ref{KLdiv} (b) and (d), as the $\sigma$ range varied from $[0.01,0.1]$ to $[0.01,1]$, the maximum of KL-divergence increases from 80 to $10^4$, which increased two orders of magnitude. 

Theorem~\ref{the1} discussed that when the asymmetry of KL-divergence is large enough, it may misguide the surrogate improvement and affect the learning performance. Next, to display the scale of  $\Delta { D_{\rm KL}}$, we visualize it in Fig.~\ref{KLdiv2} .  In the figures, the $x$ and $y$ axes represent the $\sigma_1$ and $\sigma_2$, respectively, and the different colors represent the $\Delta D_{\rm KL}$ with the same value range. Overall, the maximum of $\Delta D_{\rm KL}$ is positively related to a difference of $\mu_1$ and $\mu_2$. Meanwhile,  as the difference of $\sigma_1$ and $\sigma_2$ increase, the $\Delta D_{\rm KL}$ is increase. In Fig.\ref{KLdiv2} (c) and (d), the $\Delta D_{\rm KL}$ even increased till $10^4$, which is greater than the reward function in most RL environments. Moreover, the KL divergence is unbounded~\citep{KLd}, so the $\Delta { D_{\rm KL}}$ can be unlimited. 

\begin{figure*}[htbp]
\centering
\subfloat[]{\includegraphics[width=0.33\textwidth]{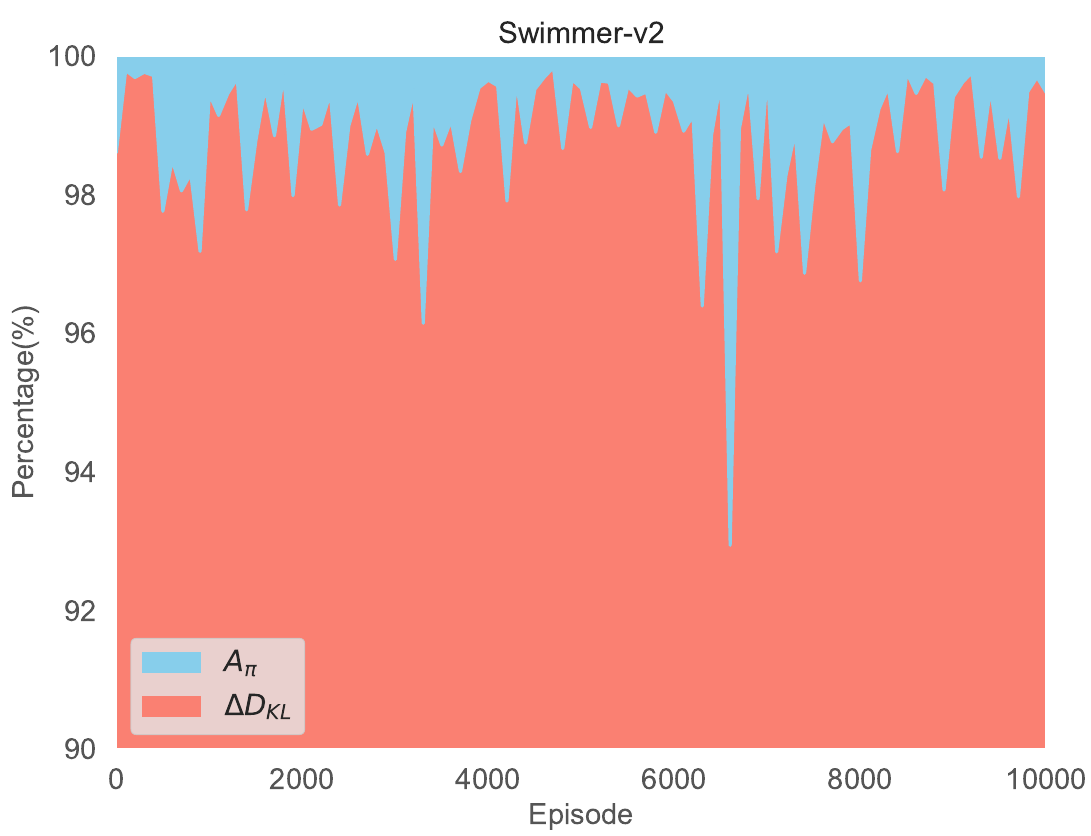}}
\subfloat[]{\includegraphics[width=0.33\textwidth]{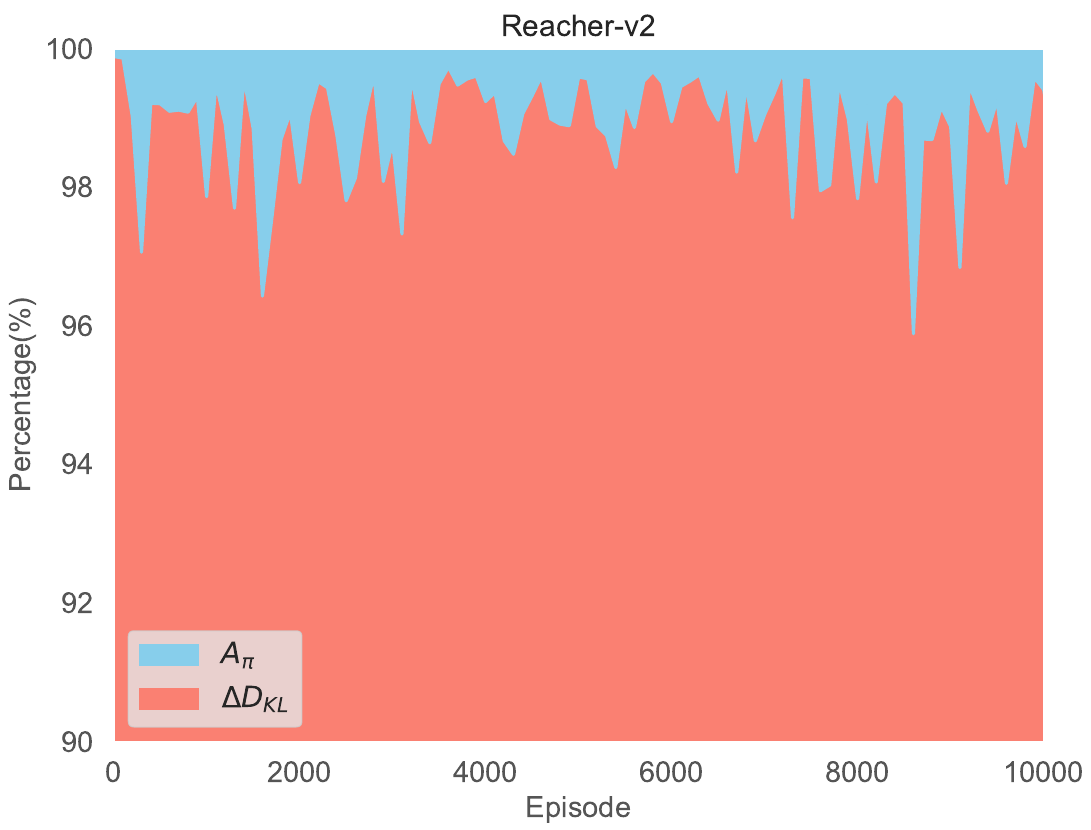}}
\subfloat[]{\includegraphics[width=0.33\textwidth]{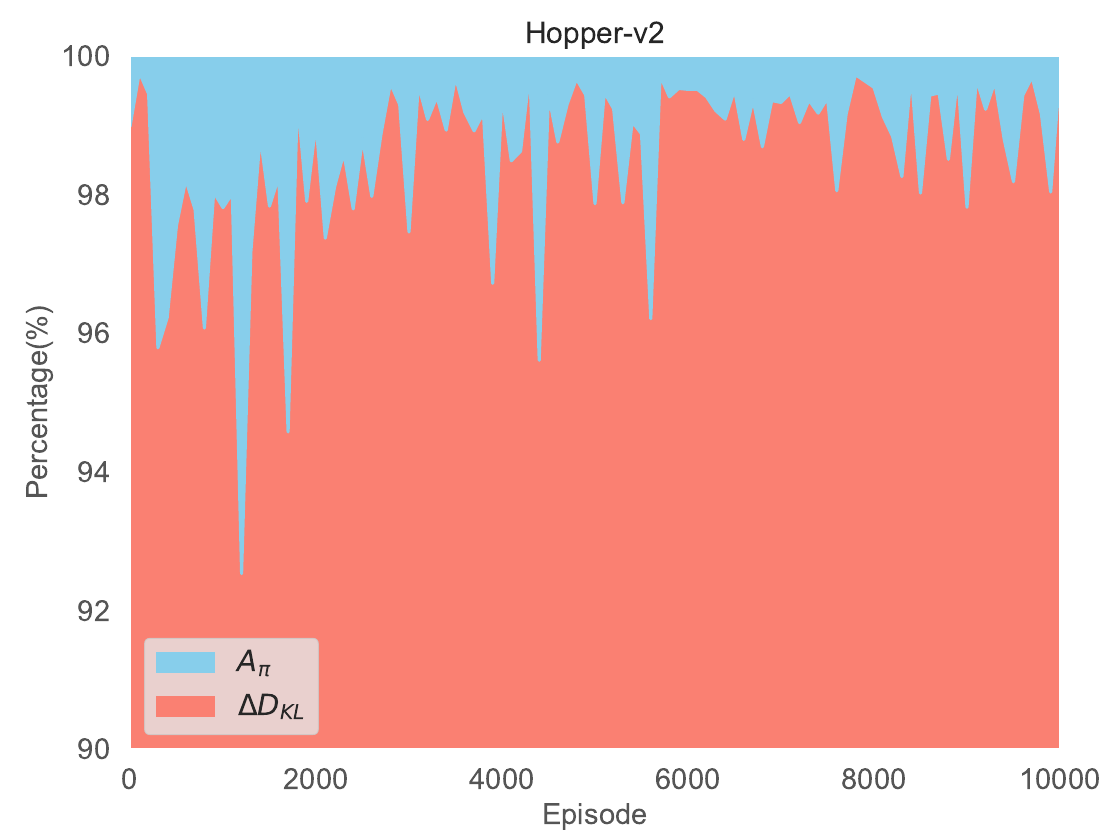}}\\
\subfloat[]{\includegraphics[width=0.33\textwidth]{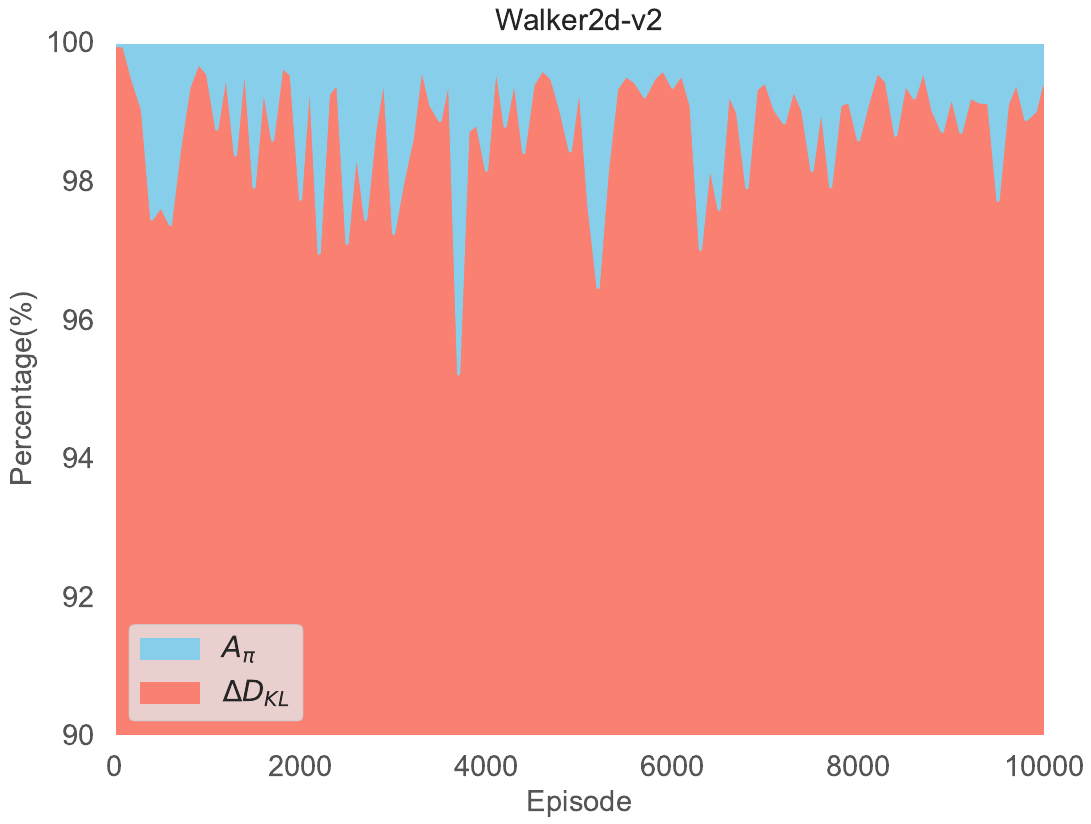}}
\subfloat[]{\includegraphics[width=0.33\textwidth]{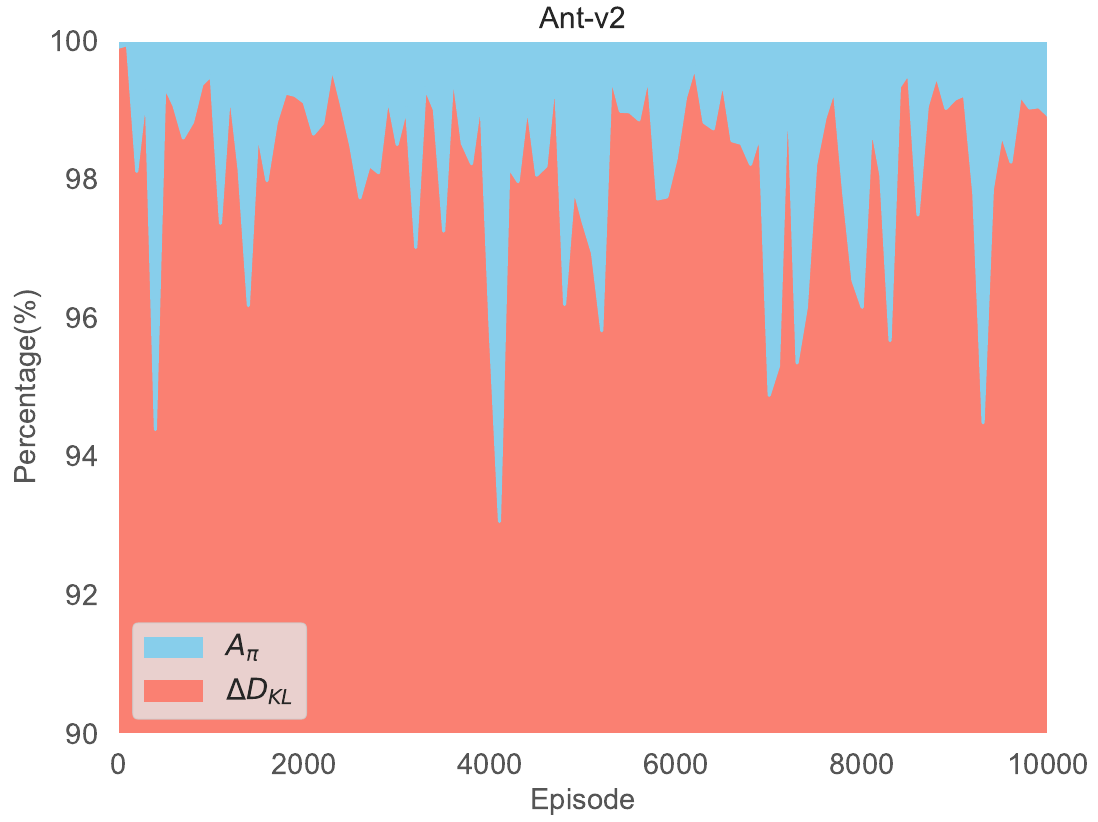}}
\subfloat[]{\includegraphics[width=0.33\textwidth]{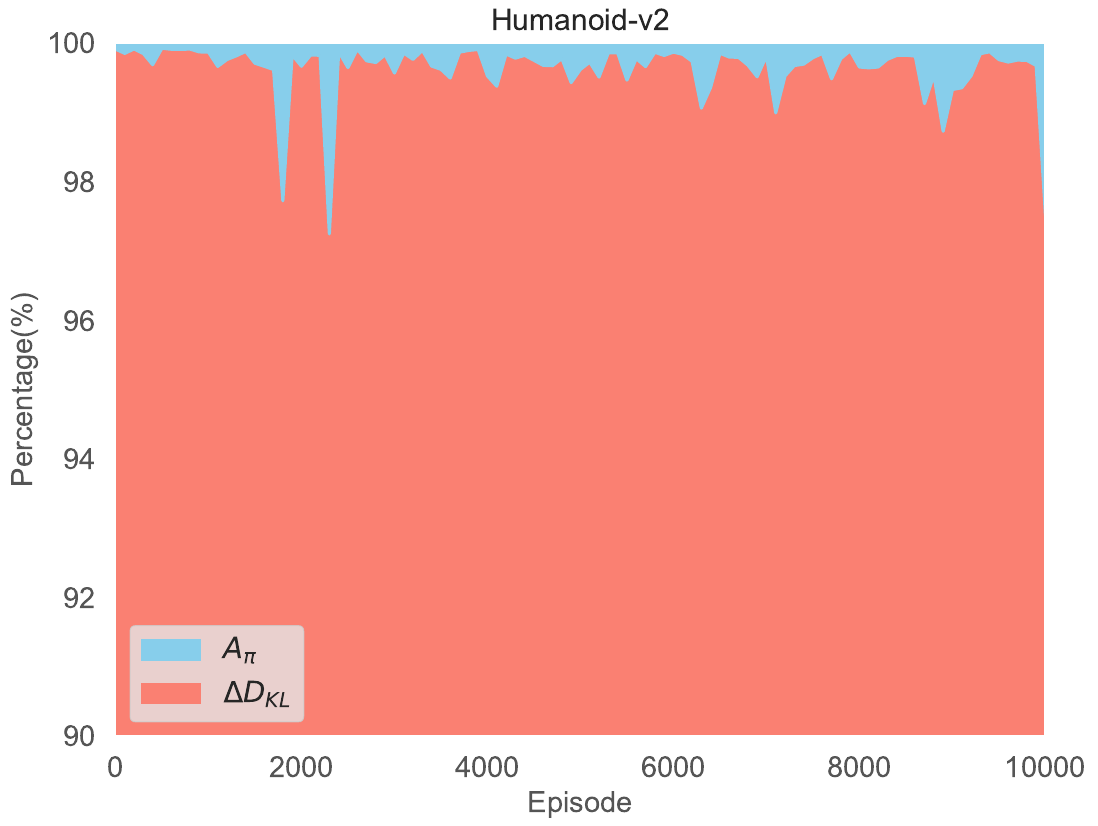}}
\caption{The scale compares between the asymmetry of KL-divergence and estimated advantage function: (a) Swimmer-v2; (b) Reacher-v2; (c) Hopper-v2; (d) Walker2d-v2; (e) Ant-v2;
(f) Humanoid-v2}
\label{KLdiv3}
\end{figure*}

In addition, to show the scale of $\Delta D_{\rm KL}$ in the real DRL scenarios, we trained the PPO-KL with 10,000 episodes, each episode with 120 maximum time steps, in the six Mujoco continuous-action tasks. We collected the absolute values of the estimated value function $\mathbb{E}[\frac{\boldsymbol{\pi}_{\boldsymbol{\theta}}}{\boldsymbol{\pi}_{\boldsymbol{\theta}_{\rm old}}}A^\pi]$ and the asymmetry of KL-divergence $\Delta D_{\rm KL}(\boldsymbol{\pi}_{\boldsymbol{\theta}_{\rm old}},\boldsymbol{\pi}_{\boldsymbol{\theta}})$ during training process, and compared their scale in Fig.~\ref{KLdiv3}. Specifically, the red area represents the percentage of $\|\Delta D_{\rm KL}(\boldsymbol{\pi}_{\boldsymbol{\theta}_{\rm old}},\boldsymbol{\pi}_{\boldsymbol{\theta}})\|$ in $\|\mathbb{E}[\frac{\boldsymbol{\pi}_{\boldsymbol{\theta}}}{\boldsymbol{\pi}_{\boldsymbol{\theta}_{\rm old}}}A^\pi]\|+\|\Delta D_{\rm KL}(\boldsymbol{\pi}_{\boldsymbol{\theta}_{\rm old}},\boldsymbol{\pi}_{\boldsymbol{\theta}})\|$ and the blue area represents the corresponding percentage of  $\|\mathbb{E}[\frac{\boldsymbol{\pi}_{\boldsymbol{\theta}}}{\boldsymbol{\pi}_{\boldsymbol{\theta}_{\rm old}}}A^\pi]\|$.
 
Because these tasks' rewards are monotonic in most training stages (the episode rewards of the training process will be shown in the next section), the improvement of the estimated value function during the training will be less than the estimated value function. Therefore, Fig.~\ref{KLdiv3} can represent the relationship between the improvement of the value function and the asymmetry of KL divergence. Overall, Fig.~\ref{KLdiv3} shows that the asymmetry of KL divergence $\Delta D_{\rm KL}$ occupied over 90\% of the total, which is significantly larger than the corresponding estimated value function in all the six mujoco environments. In sum, according to the theorem~\ref{the1}, the surrogate of the PPO-KL can not guide the policy improvement due to the significant $\Delta D_{\rm KL}$ in the six environments. 

Specially, we select the results of Hopper-v2, Walker2d-v2, and Humanoid-v2 (Fig.~\ref{KLdiv3} (c), (d), (f)) to explain because these tasks' rewards are in the same order of magnitude. Theorem~\ref{the1} pointed out that the asymmetry of KL divergence is positively related to the number of actions. Combine the number of actions of the tasks that show in Tab.~\ref{envtab}, the results show the percentages of the $\Delta D_{\rm KL}$ increase as the number of the actions increases, which validates this sub-conclusion.

\subsection{Experimental Results}
\begin{figure*}[htbp]
\centering
\subfloat[]{\includegraphics[width=0.33\textwidth]{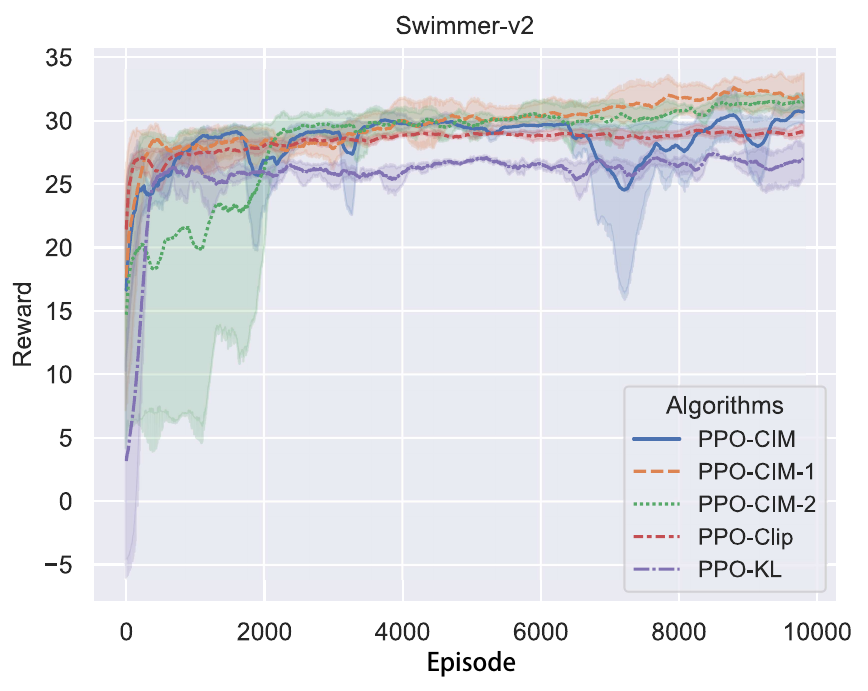}}
\subfloat[]{\includegraphics[width=0.33\textwidth]{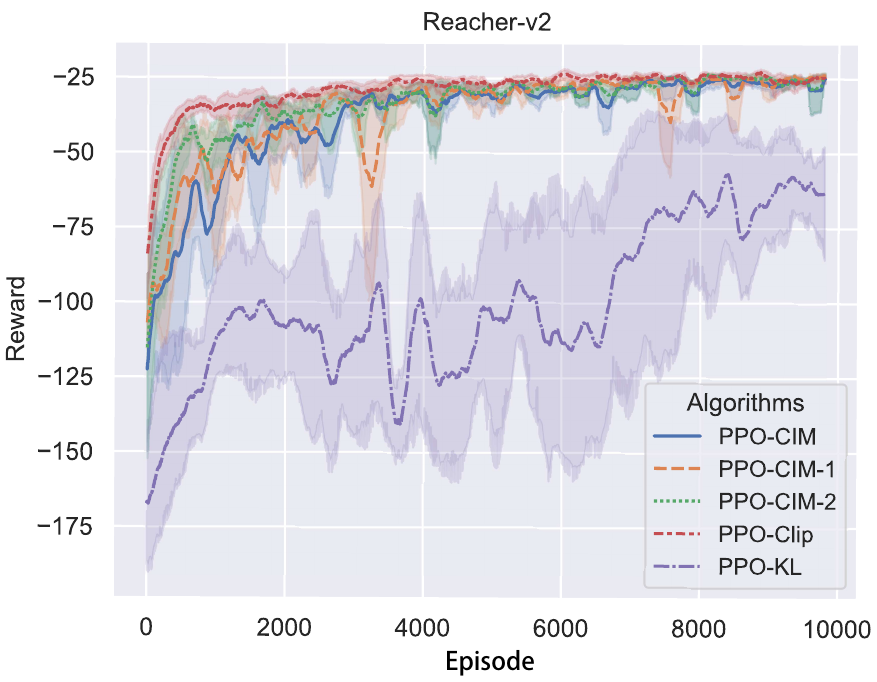}}
\subfloat[]{\includegraphics[width=0.33\textwidth]{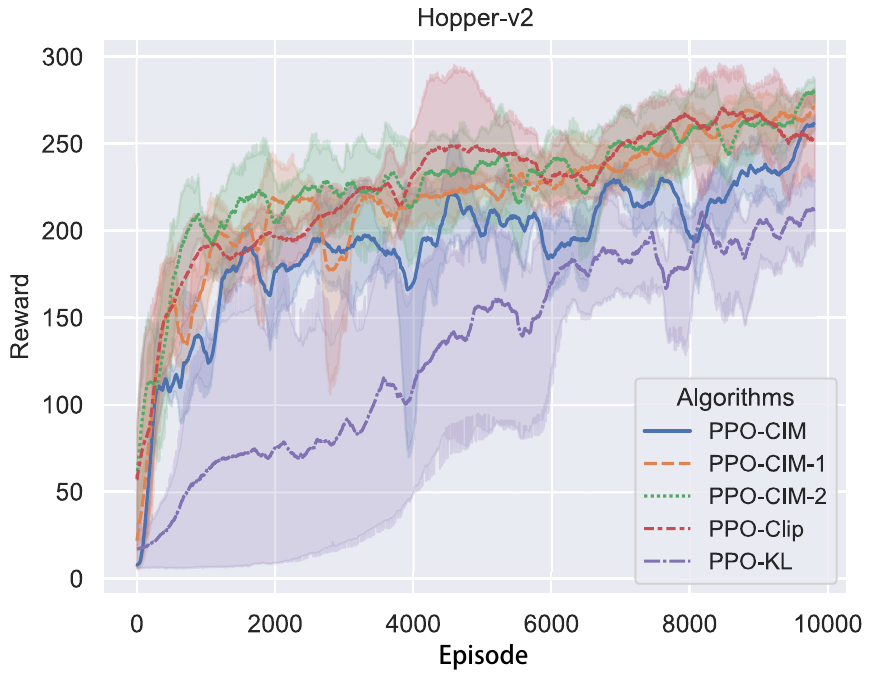}}\\
\subfloat[]{\includegraphics[width=0.33\textwidth]{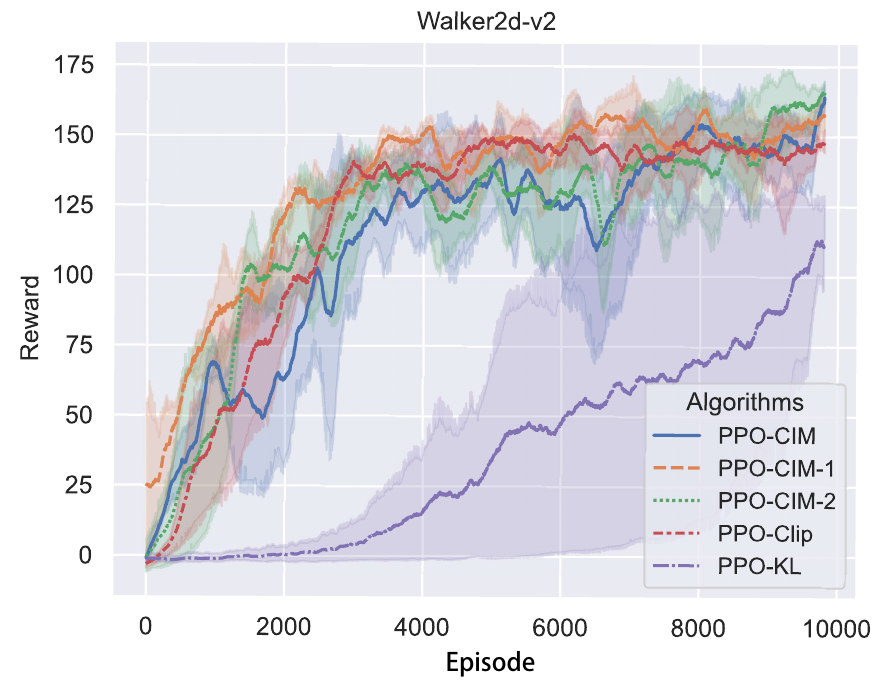}}
\subfloat[]{\includegraphics[width=0.33\textwidth]{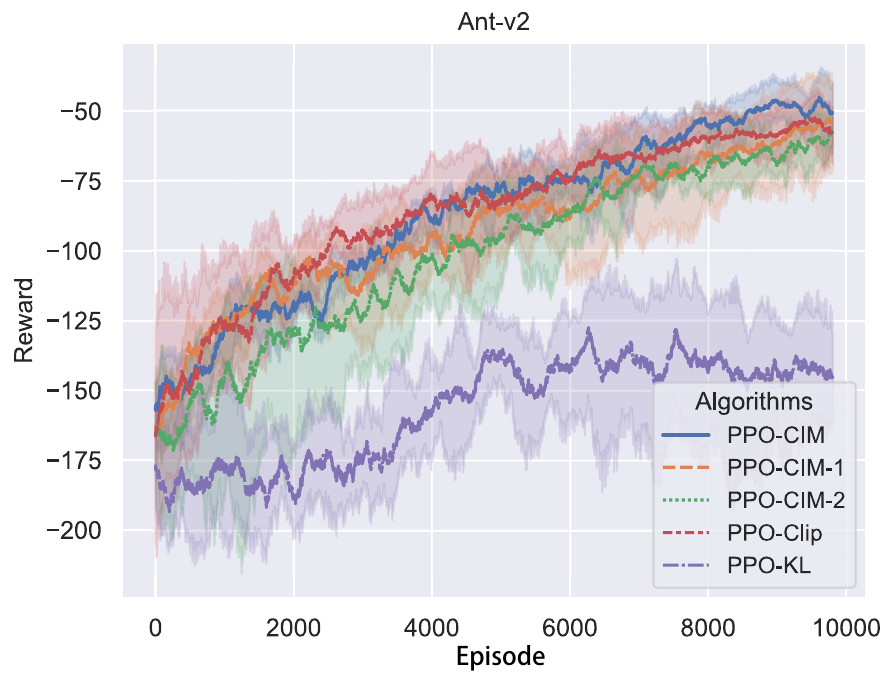}}
\subfloat[]{\includegraphics[width=0.33\textwidth]{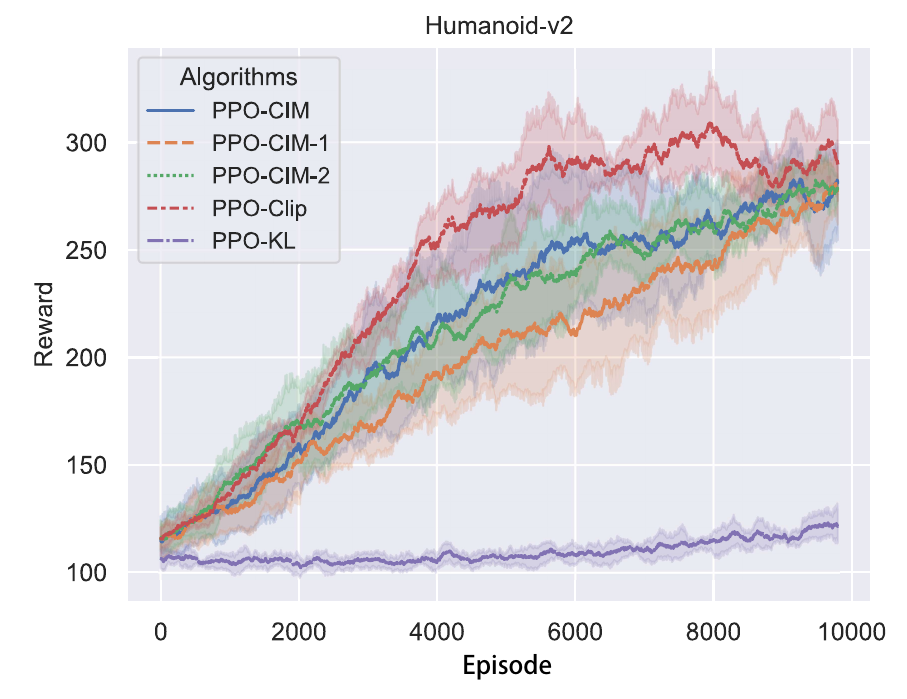}}
\caption{The training results in the Mujoco tasks: (a) Swimmer-v2; (b) Reacher-v2; (c) Hopper-v2; (d) Walker2d-v2; (e) Ant-v2; (e) Humanoid-v2}
\label{result}
\end{figure*}

To evaluate the performance of PPO-CIM, we train the PPO-KL, PPO-Clip, PPO-CIM, PPO-CIM-1, and PPO-CIM-2 in the six Mujoco continuous-action tasks with 10,000 episodes. In the tasks, the maximum time-step length of each episode is set as 120, and each algorithm was repeated with different random seeds four times. The episode will be halted when the maximum length is reached, or the task is done, and then the reward during the episode will be added as the episode reward. After training, we visualize the learning curves in Fig.~\ref{result}.  
In Fig.~\ref{result}, the colorful lines and shades represent the average episode rewards and their standard variance of all four repeated experiments, respectively.

\begin{enumerate}[(1)]
    \item As the action number increases, the performance of PPO-KL decreases. Specifically, the learning curve of PPO-KL is slightly lower than the other algorithms in Swimmer-v2 (action number is 2). However, in the subsequent tasks, the learning curves are significantly lower than the others, especially in Ant-v2 (action number 111) and Humanoid-v2 (action number 376). Combined with the previous experiment results on the asymmetry of KL divergence, the training results validate that when the action number increases, the asymmetry of KL will increase and affect the surrogate.

    \item The proposed algorithms PPO-CIM, PPO-CIM-1, and PPO-CIM-2 perform better than PPO-KL significantly, the average reward of the proposed algorithms is higher than PPO-KL, and the standard variance of the proposed algorithms is lower than the PPO-KL. Moreover, in all tasks except Humanoid-v2, the proposed algorithms are slightly better than PPO-Clip.

    \item Compared with PPO-CIM, the approximation methods PPO-CIM-1 and PPO-CIM-2 are more stable. Specifically, when the action number is small (such as Swimmer-v2, Reacher-v2, and Hopper-v2), the average reward is smoother and their stand variance is less than the corresponding PPO-CIM. As the action number increases, PPO-CIM's performance and stability reach the same level as PPO-CIM-1 and PPO-CIM-2, and even better than them.
\end{enumerate}
\begin{table*}[t]
\centering
\caption{The average reward in 100 episodes of Mujoco tasks.}
\begin{tabular}{ccccccc}
\hline
                              & \multicolumn{1}{c}{Swimmer} & \multicolumn{1}{c}{Reacher} & Hopper & \multicolumn{1}{c}{Walker2d} & \multicolumn{1}{c}{Ant} & Humanoid\\ \hline
PPO-KL                        & \multicolumn{1}{c}{$27.034\pm2.61$}           & \multicolumn{1}{c}{$ -60.207\pm 14.090$}           &           $206.228\pm40.614$& \multicolumn{1}{c}{$107.219\pm49.738$}                     & \multicolumn{1}{c}{$-148.112\pm125.83$} &                $ 122.136\pm43.206$\\ 
PPO-Clip                      & \multicolumn{1}{c}{$29.025\pm 1.32$}           & \multicolumn{1}{c}{$ -24.666\pm10.393$}           &           $251.151\pm 41.175$& \multicolumn{1}{c}{$146.158\pm 42.481$}                     & \multicolumn{1}{c}{$ -58.461\pm 63.174$} &                $ {\bf 286.530\pm 81.952}$\\ 
PPO-CIM                       &                                $29.211\pm1.481$&                                $-25.158\pm 3.001$&           $ 247.751\pm 37.251$&                                          $153.859\pm41.503$&                                     ${\bf  -51.209\pm62.462}$ &$ 268.835\pm88.089$\\
\multicolumn{1}{c}{PPO-CIM-1} &                                ${\bf 30.538\pm 1.608}$&                                $-22.448\pm4.086$&           $259.584\pm34.118$&                                          $153.525\pm38.799$&                                     $ -57.490\pm54.272$ &$ 261.760\pm89.193$\\ 
\multicolumn{1}{c}{PPO-CIM-2} &                                $29.978\pm2.031$&                                $ {\bf -21.945\pm2.657}$&           ${\bf 266.117\pm37.856}$&                                          ${\bf159.984\pm35.778}$&                                     $ -54.596\pm 36.641$ &$260.723\pm 81.622$\\ \hline
\end{tabular}
\label{test}
\end{table*}

After training, we tested the trained algorithms with 100 independent episodes in the six Mujoco continuous-action tasks and displayed the average test episode rewards and their standard variances in Tab.~\ref{test}.

Tab.~\ref{test} provides the quantitative results to accurately evaluate the performance of the algorithms. Overall, in most of the tasks, the average rewards of PPO-CIM-2 are the largest and its variances are also the smallest. Compared with PPO-KL, the proposed algorithms are greatly improved in whatever performance and stability, which indicated that it is efficient to extend KL divergence from Euclidean space to RKHS.

\textbf{To a sum, although the PPO-CIM algorithms are only reach the similar level of PPO-Clip, the meaning of the PPO-CIM is significant. It indicated why PPO-KL perform bad, and the methods that extend the KL divergence from Euclidean space to RKHS is efficient. Our research provides a novel view, it inspired the researchers to utilize the kernel methods to improve PPO with theoretical substantitation.}

\section{Conclusion}

This paper mainly studied the asymmetry of KL divergence in the PPO-KL and proved it may affect the policy improvement of the algorithm. To solve this issue, we represented the KL divergence approximation with the inner product in RKHS and discovered it is the CIM in Euclidean space. Inspired by it, we proposed the PPO-CIM algorithm that extended the KL divergence to RKHS, which can guarantee the computational complexity of PPO-CIM's policy gradient is lower than the corresponding PPO-KL. Moreover, we gave the condition that associates the selection of the RKHS such that the new policy can be restricted in the trust region. In addition, we presented two approximations of PPO-CIM to further improve learning efficiency through expanding the CIM. Finally, our experiments validate that the asymmetry of KL divergence is significant enough to affect the policy improvement of PPO-KL, and the proposed PPO-CIM can perform better than the PPO-KL and PPO-Clip in six Mujoco continuous-action space tasks.


\appendix
\section{Appendix: The proof of Lemma \ref{lemma2}}\label{app2}
\begin{proof}
The KL divergence between $\boldsymbol{\pi}_1(\boldsymbol{a}|\boldsymbol{s})$ and $\boldsymbol{\pi}_2(\boldsymbol{a}|\boldsymbol{s})$ is:
\begin{align}
&\kl{\boldsymbol{\pi}_1(\boldsymbol{\cdot}|\boldsymbol{s})}{\boldsymbol{\pi}_2(\boldsymbol{\cdot}|\boldsymbol{s})}\nonumber \\
=&\int_{\boldsymbol{a}} \boldsymbol{\pi}_1(\boldsymbol{a}|\boldsymbol{s})^\top\log\frac{\boldsymbol{\pi}_1(\boldsymbol{a}|\boldsymbol{s})}{\boldsymbol{\pi}_2(\boldsymbol{a}|\boldsymbol{s})}d\boldsymbol{a} \nonumber \\
=&-\int_{\boldsymbol{a}} \boldsymbol{\pi}_1(\boldsymbol{a}|\boldsymbol{s})^\top\log[\boldsymbol{\pi}_2(\boldsymbol{a}|\boldsymbol{s})] d\boldsymbol{a} \nonumber \\
&+\int_{\boldsymbol{a}} \boldsymbol{\pi}_1(\boldsymbol{a}|\boldsymbol{s})^\top\log[\boldsymbol{\pi}_2(\boldsymbol{a}|\boldsymbol{s})]d\boldsymbol{a}.\label{rst3}
\end{align}
Considering the second part of Eq. (\ref{rst3}), expanding $\boldsymbol{\pi}_2(\boldsymbol{a}|\boldsymbol{s})$, we have:

\begin{align}
&\int_{\boldsymbol{a}} \boldsymbol{\pi}_1(\boldsymbol{a}|\boldsymbol{s})^\top\log[\boldsymbol{\pi}_2(\boldsymbol{a}|\boldsymbol{s})] d\boldsymbol{a} \nonumber \\
=&\int_{\boldsymbol{a}} \boldsymbol{\pi}_1(\boldsymbol{a}|\boldsymbol{s})^\top\log[(2\pi)^{-\frac{n}{2}}|\mathbf{F}_{\theta_2}^{\boldsymbol{\Sigma}}|^{\frac{1}{2}}\exp\{-\frac{1}{2}(\boldsymbol{a}-\mathbf{F}_{\theta_2}^{\boldsymbol{\mu}})^\top\nonumber \\
&\cdot(\mathbf{F}_{\theta_2}^{\boldsymbol{\Sigma}})^{-1} 
(\boldsymbol{a}-\mathbf{F}_{\theta_2}^{\boldsymbol{\mu}})\}] d\boldsymbol{a} \nonumber \\
=&\int_{\boldsymbol{a}} \boldsymbol{\pi}_1(\boldsymbol{a}|\boldsymbol{s})^\top(-\frac{n}{2}\log2\pi-\frac{1}{2}\log|\mathbf{F}_{\theta_2}^{\boldsymbol{\Sigma}}|)d\boldsymbol{a} \nonumber\\
&-\frac{1}{2}\int_{\boldsymbol{a}} (\boldsymbol{a}-\mathbf{F}_{\theta_2}^{\boldsymbol{\mu}})^\top(\mathbf{F}_{\theta_2}^{\boldsymbol{\Sigma}})^{-1}(\boldsymbol{a}-\mathbf{F}_{\theta_2}^{\boldsymbol{\mu}})\boldsymbol{\pi}_1(\boldsymbol{a}|\boldsymbol{s})d\boldsymbol{a} \nonumber \\
=& -\frac{1}{2}\int_{\boldsymbol{a}} (\boldsymbol{a}-\mathbf{F}_{\theta_2}^{\boldsymbol{\mu}})^\top(\mathbf{F}_{\theta_2}^{\boldsymbol{\Sigma}})^{-1}(\boldsymbol{a}-\mathbf{F}_{\theta_2}^{\boldsymbol{\mu}})\boldsymbol{\pi}_1(\boldsymbol{a}|\boldsymbol{s})d\boldsymbol{a} \nonumber \\
&-\frac{n}{2}\log2\pi-\frac{1}{2}\log|\mathbf{F}_{\theta_2}^{\boldsymbol{\Sigma}}| \nonumber
\end{align}

Because ${\rm dim}\big[(\boldsymbol{a}-\mathbf{F}_{\theta_2}^{\boldsymbol{\mu}})^\top(\mathbf{F}_{\theta_2}^{\boldsymbol{\Sigma}})^{-1}(\boldsymbol{a}-\mathbf{F}_{\theta_2}^{\boldsymbol{\mu}})\big]=1$, we have ${\rm Trace}[(\boldsymbol{a}-\mathbf{F}_{\theta_2}^{\boldsymbol{\mu}})^\top(\mathbf{F}_{\theta_2}^{\boldsymbol{\Sigma}})^{-1}(\boldsymbol{a}-\mathbf{F}_{\theta_2}^{\boldsymbol{\mu}})]=(\boldsymbol{a}-\mathbf{F}_{\theta_2}^{\boldsymbol{\mu}})^\top(\mathbf{F}_{\theta_2}^{\boldsymbol{\Sigma}})^{-1}(\boldsymbol{a}-\mathbf{F}_{\theta_2}^{\boldsymbol{\mu}})$. Then we have:

\begin{align}
&\frac{1}{2}\int_{\boldsymbol{a}} (\boldsymbol{a}-\mathbf{F}_{\theta_2}^{\boldsymbol{\mu}})^{\top}(\mathbf{F}_{\theta_2}^{\boldsymbol{\Sigma}})^{-1}(\boldsymbol{a}-\mathbf{F}_{\theta_2}^{\boldsymbol{\mu}})\boldsymbol{\pi}_1(\boldsymbol{a}|\boldsymbol{s})d\boldsymbol{a} \nonumber \\
=&\frac{1}{2}\int_{\boldsymbol{a}} \rm{Trace}[(\boldsymbol{a}-\mathbf{F}_{\theta_2}^{\boldsymbol{\mu}})^{\top}(\mathbf{F}_{\theta_2}^{\boldsymbol{\Sigma}})^{-1}(\boldsymbol{a}-\mathbf{F}_{\theta_2}^{\boldsymbol{\mu}})\boldsymbol{\pi}_1(\boldsymbol{a}|\boldsymbol{s})]d\boldsymbol{a}\nonumber \\
=&\frac{1}{2}\int_{\boldsymbol{a}} \rm{Trace}[(\boldsymbol{a}-\mathbf{F}_{\theta_2}^{\boldsymbol{\mu}})(\boldsymbol{a}-\mathbf{F}_{\theta_2}^{\boldsymbol{\mu}})^{\top}(\mathbf{F}_{\theta_2}^{\boldsymbol{\Sigma}})^{-1}\boldsymbol{\pi}_1(\boldsymbol{a}|\boldsymbol{s})]d\boldsymbol{a}\nonumber \\
=&\frac{1}{2}\int_{\boldsymbol{a}} \rm{Trace}\{[(\boldsymbol{a}-\mathbf{F}_{\theta_1}^{\boldsymbol{\mu}})+(\mathbf{F}_{\theta_1}^{\boldsymbol{\mu}}-\mathbf{F}_{\theta_2}^{\boldsymbol{\mu}})][(\boldsymbol{a}-\mathbf{F}_{\theta_1}^{\boldsymbol{\mu}})\nonumber\\
&+(\mathbf{F}_{\theta_1}^{\boldsymbol{\mu}}-\mathbf{F}_{\theta_2}^{\boldsymbol{\mu}})]^{\top}(\mathbf{F}_{\theta_2}^{\boldsymbol{\Sigma}})^{-1}\boldsymbol{\pi}_1(\boldsymbol{a}|\boldsymbol{s})\}d\boldsymbol{a} \nonumber \\
=&\frac{1}{2}\bigg\{\int_{\boldsymbol{a}} \rm{Trace}[(\boldsymbol{a}-\mathbf{F}_{\theta_1}^{\boldsymbol{\mu}})(\boldsymbol{a}-\mathbf{F}_{\theta_1}^{\boldsymbol{\mu}})^{\top}(\mathbf{F}_{\theta_2}^{\boldsymbol{\Sigma}})^{-1}\boldsymbol{\pi}_1(\boldsymbol{a}|\boldsymbol{s})]d\boldsymbol{a} \nonumber \\
&-2\int_{\boldsymbol{a}} \rm{Trace}[(\mathbf{F}_{\theta_2}^{\boldsymbol{\mu}}-\mathbf{F}_{\theta_1}^{\boldsymbol{\mu}})(\boldsymbol{a}-\mathbf{F}_{\theta_2}^{\boldsymbol{\mu}})^{\top}(\mathbf{F}_{\theta_2}^{\boldsymbol{\Sigma}})^{-1}\boldsymbol{\pi}_1(\boldsymbol{a}|\boldsymbol{s})]d\boldsymbol{a} \nonumber \\
&+\int_{\boldsymbol{a}} \rm{Trace}[(\mathbf{F}_{\theta_2}^{\boldsymbol{\Sigma}})^{-1}(\mathbf{F}_{\theta_2}^{\boldsymbol{\mu}}-\mathbf{F}_{\theta_1}^{\boldsymbol{\mu}})(\mathbf{F}_{\theta_2}^{\boldsymbol{\mu}}-\mathbf{F}_{\theta_1}^{\boldsymbol{\mu}})^\top \boldsymbol{\pi}_1(\boldsymbol{a}|\boldsymbol{s})]d\boldsymbol{a} \bigg\} \nonumber \\
=&-\frac{1}{2}\rm{Trace}((\mathbf{F}_{\theta_2}^{\boldsymbol{\Sigma}})^{-1}(\mathbf{F}_{\theta_2}^{\boldsymbol{\mu}}-\mathbf{F}_{\theta_1}^{\boldsymbol{\mu}})(\mathbf{F}_{\theta_2}^{\boldsymbol{\mu}}-\mathbf{F}_{\theta_1}^{\boldsymbol{\mu}})^\top) \nonumber\\
&-\frac{1}{2}\rm{Trace}[(\mathbf{F}_{\theta_2}^{\boldsymbol{\Sigma}})^{-1}\mathbf{F}_{\theta_1}^{\boldsymbol{\Sigma}}]
\end{align}

Therefore, we obtain the expression as follows:
\begin{align}
&\int_{\boldsymbol{a}} \boldsymbol{\pi}_1(\boldsymbol{a}|\boldsymbol{s})^\top\log[\boldsymbol{\pi}_2(\boldsymbol{a}|\boldsymbol{s})] d\boldsymbol{a} \nonumber \\
=&\frac{1}{2}\rm{Trace}[(\mathbf{F}_{\theta_2}^{\boldsymbol{\Sigma}})^{-1}(\mathbf{F}_{\theta_2}^{\boldsymbol{\mu}}-\mathbf{F}_{\theta_1}^{\boldsymbol{\mu}})(\mathbf{F}_{\theta_2}^{\boldsymbol{\mu}}-\mathbf{F}_{\theta_1}^{\boldsymbol{\mu}})^\top]\nonumber \\
&+\frac{1}{2}\rm{Trace}[(\mathbf{F}_{\theta_2}^{\boldsymbol{\Sigma}})^{-1}\mathbf{F}_{\theta_1}^{\boldsymbol{\Sigma}}]-\frac{n}{2}\log2\pi-\frac{1}{2}\log|\mathbf{F}_{\theta_2}^{\boldsymbol{\Sigma}}|\nonumber \\
\end{align}

By the similar way, we have:

\begin{align}
&\int_{\boldsymbol{a}} \boldsymbol{\pi}_1(\boldsymbol{a}|\boldsymbol{s})^\top\log[\boldsymbol{\pi}_1(\boldsymbol{a}|\boldsymbol{s})] d\boldsymbol{a} \nonumber \\
=&-\frac{n}{2}\log2\pi-\frac{1}{2}\log|\mathbf{F}_{\theta_1}^{\boldsymbol{\Sigma}}|-\frac{n}{2}\nonumber 
\end{align}

Finally, we have the KL divergence of multi-dimensional normal distribution :
\begin{align}
&\kl{\boldsymbol{\pi}_1(\boldsymbol{\cdot}|\boldsymbol{s})}{\boldsymbol{\pi}_2(\boldsymbol{\cdot}|\boldsymbol{s})}\nonumber \\
=&\int_{\boldsymbol{a}} \boldsymbol{\pi}_1(\boldsymbol{a}|\boldsymbol{s})^\top\log\frac{\boldsymbol{\pi}_1(\boldsymbol{a}|\boldsymbol{s})}{\boldsymbol{\pi}_2(\boldsymbol{a}|\boldsymbol{s})}d\boldsymbol{a} \nonumber \\
=&-\int_{\boldsymbol{a}} \boldsymbol{\pi}_1(\boldsymbol{a}|\boldsymbol{s})^\top\log[\boldsymbol{\pi}_2(\boldsymbol{a}|\boldsymbol{s})] d\boldsymbol{a}+\int_{\boldsymbol{a}} \boldsymbol{\pi}_1(\boldsymbol{a}|\boldsymbol{s})^\top
\nonumber \\
&\cdot\log[\boldsymbol{\pi}_1(\boldsymbol{a}|\boldsymbol{s})]d\boldsymbol{a}\nonumber \\
=&\frac{1}{2}\rm{Trace}[(\mathbf{F}_{\theta_1}^{\boldsymbol{\Sigma}})^{-1}\mathbf{F}_{\theta_2}^{\boldsymbol{\Sigma}}+(\mathbf{F}_{\theta_2}^{\boldsymbol{\Sigma}})^{-1}(\mathbf{F}_{\theta_2}^{\boldsymbol{\mu}}-\mathbf{F}_{\theta_1}^{\boldsymbol{\mu}})(\mathbf{F}_{\theta_2}^{\boldsymbol{\mu}}-\mathbf{F}_{\theta_1}^{\boldsymbol{\mu}})^\top] \nonumber\\
&-\frac{n}{2}+\frac{1}{2}\log\frac{|\mathbf{F}_{\theta_1}^{\boldsymbol{\Sigma}}|}{|\mathbf{F}_{\theta_2}^{\boldsymbol{\Sigma}}|}
\end{align}

Therefore, the asymmetry of KL divergence can be represented as follows:
\begin{align}
&\Delta D_{\rm KL}(\boldsymbol{\pi}_1(\boldsymbol{\cdot}|\boldsymbol{s}),\boldsymbol{\pi}_2(\boldsymbol{\cdot}|\boldsymbol{s}))\nonumber\\
=&\kl{\boldsymbol{\pi}_1(\boldsymbol{\cdot}|\boldsymbol{s})}{\boldsymbol{\pi}_2(\boldsymbol{\cdot}|\boldsymbol{s})}-\kl{\boldsymbol{\pi}_2(\boldsymbol{\cdot}|\boldsymbol{s})}{\boldsymbol{\pi}_1(\boldsymbol{\cdot}|\boldsymbol{s})}\nonumber \\
=&\frac{1}{2}\big\{\rm{Trace}[(\mathbf{F}_{\theta_1}^{\boldsymbol{\Sigma}})^{-1}\mathbf{F}_{\theta_2}^{\boldsymbol{\Sigma}}+(\mathbf{F}_{\theta_2}^{\boldsymbol{\Sigma}})^{-1}(\mathbf{F}_{\theta_2}^{\boldsymbol{\mu}}-\mathbf{F}_{\theta_1}^{\boldsymbol{\mu}})(\mathbf{F}_{\theta_2}^{\boldsymbol{\mu}}-\mathbf{F}_{\theta_1}^{\boldsymbol{\mu}})^\top]\nonumber \\
&-\rm{Trace}[(\mathbf{F}_{\theta_2}^{\boldsymbol{\Sigma}})^{-1}\mathbf{F}_{\theta_1}^{\boldsymbol{\Sigma}}+(\mathbf{F}_{\theta_1}^{\boldsymbol{\Sigma}})^{-1}(\mathbf{F}_{\theta_1}^{\boldsymbol{\mu}}-\mathbf{F}_{\theta_2}^{\boldsymbol{\mu}})(\mathbf{F}_{\theta_1}^{\boldsymbol{\mu}}-\mathbf{F}_{\theta_2}^{\boldsymbol{\mu}})^\top]\big\} \nonumber\\
&+\log\frac{|\mathbf{F}_{\theta_1}^{\boldsymbol{\Sigma}}|}{|\mathbf{F}_{\theta_2}^{\boldsymbol{\Sigma}}|}\nonumber
\end{align}
\end{proof}

\bibliographystyle{cas-model2-names}
\bibliography{Ref}

\end{document}